\documentclass{article}

\usepackage{microtype}
\usepackage{graphicx}
\usepackage{subcaption}
\usepackage{booktabs} 
\usepackage{tcolorbox}

\usepackage{hyperref}
\usepackage{enumitem}

\usepackage[preprint]{icml2026}

\usepackage{amsmath}
\usepackage{amssymb}
\usepackage{mathtools}
\usepackage{amsthm}

\usepackage[capitalize,noabbrev]{cleveref}

\theoremstyle{plain}

\theoremstyle{definition}

\theoremstyle{remark}

\usepackage[textsize=tiny]{todonotes}

\icmltitlerunning{TeamMedAgents: Pareto-Efficient Multi-Agent Medical Reasoning Through Teamwork Theory}

\begin{document}

\twocolumn[
  \icmltitle{TeamMedAgents: Pareto-Efficient Multi-Agent Medical Reasoning Through Teamwork Theory}

  \icmlsetsymbol{equal}{*}

  \begin{icmlauthorlist}
    \icmlauthor{Pranav Mishra}{uic}
    \icmlauthor{Mohammad Arvan}{uic}
    \icmlauthor{Mohan Zalake}{uic}
  \end{icmlauthorlist}

  \icmlaffiliation{uic}{University of Illinois Chicago, IL, USA}

  \icmlcorrespondingauthor{Pranav Mishra}{pmishr32@uic.edu}
  \icmlcorrespondingauthor{Mohammad Arvan}{marvan2@uic.edu}
  \icmlcorrespondingauthor{Mohan Zalake}{mzalak2@uic.edu}

  \icmlkeywords{Machine Learning, ICML}

  \vskip 0.3in
]

\printAffiliationsAndNotice{}  

\begin{abstract}
Complex medical reasoning has historically required frontier language models to achieve clinically-acceptable accuracy, creating computational barriers that limit deployment in resource-constrained clinical settings. We present TeamMedAgents, a modular multi-agent framework that translates Salas et al.'s evidence-based teamwork theory into computational mechanisms—shared mental models, team leadership, team orientation, trust networks, and mutual monitoring—enabling Small Language Models to perform multi-step clinical reasoning efficiently. Evaluation across 8 medical benchmarks demonstrates that TeamMedAgents advances the Pareto efficiency frontier by 1–2 orders of magnitude, achieving competitive accuracy at substantially lower token cost than MDAgents, MedAgents, DyLAN, and ReConcile. The framework exhibits the lowest cross-dataset variance among multi-agent approaches, enabling deployment without per-task tuning. Our results establish that theory-grounded coordination mechanisms provide essential scaffolding for deploying efficient medical AI in resource-constrained clinical environments. 

\end{abstract}

\section{Introduction}

Clinical decision-making demands integration of specialized medical knowledge, systematic reasoning under uncertainty, and coordination across diagnostic perspectives \citep{salas2005big}. Large language models demonstrate proficiency on medical knowledge tasks, yet deployment faces critical barriers: computational cost, inference latency, and resource requirements limiting point-of-care accessibility \citep{kim2024mdagents}. While frontier models achieve high accuracy, operational demands prove prohibitive for real-world integration. Recent advances in inference-time scaling demonstrate strategic test-time compute allocation yields substantial gains \citep{snell2024scaling_test_time,balachandran2025inference_time}, yet remain largely unexplored for enabling efficient medical reasoning with resource-constrained models.

Small Language Models (1-7B parameters) offer compelling deployment advantages: reduced latency, lower computational requirements, privacy-preserving local execution \citep{wu2024small_medical,han2024opensource_slm,gemma2025technical}. Architectural innovations, knowledge distillation, and domain-specific training enable competitive performance on focused tasks \citep{zhang2025small_punch}. However, complex clinical reasoning—differential diagnosis, multi-modal evidence integration, multi-step inference—remains challenging. Zero-shot and few-shot SLM approaches show substantial deficits (typically 40-50\% accuracy), particularly on systematic deliberation tasks \citep{liu2024small_measurements}, limiting clinical applicability where reasoning quality impacts patient safety.

A critical gap exists in translating SLM efficiency benefits into viable clinical systems. While Chain-of-Thought prompting and knowledge distillation improve individual performance \citep{wei2022cot_reasoning,shridhar2023symbolic_cot}, gains remain limited when complexity exceeds capacity. Multi-agent frameworks achieve superior results through specialist coordination \citep{kim2024mdagents,tang2024medagents,liu2024dylan,chen2024reconcile}, yet focus on frontier models without investigating whether structured teamwork scaffolds smaller models. Cost-optimal inference work \citep{chen2023frugalgpt,erol2025cost_of_pass} demonstrates Pareto-efficient selection feasibility but does not address whether \textit{structured coordination} advances efficiency frontiers beyond single-agent approaches. Bridging this requires principled methods translating collaboration patterns into computational mechanisms compensating for agent limitations while maintaining practical costs.

\textbf{Our Approach:} We present TeamMedAgents, a modular framework systematically integrating organizational psychology teamwork principles for efficient medical reasoning with Small Language Models. Building on Salas et al.'s "Big Five" \citep{salas2005big}, we operationalize five independently configurable components: team leadership (structured oversight), shared mental models (diagnostic context), team orientation (hierarchical specialist roles), mutual trust (reasoning quality weighting), mutual monitoring (peer review with feedback). Four-phase deliberation with dynamic recruitment (2-4 specialists) translates validated teamwork patterns into computational scaffolding for systematic clinical reasoning.

We validate across two scales: (1) Small Language Models (Gemma-3-4B, MedGemma-4B) demonstrate efficient reasoning at practical speeds; (2) frontier models (GPT-4o) validate principles enhance performance regardless of size. Evaluation across eight medical benchmarks reveals Gemma-3-4B with TeamMedAgents achieves 49.4\% average accuracy using single fixed configuration (TMA-AllCompon)—outperforming competing frameworks (MedAgents 48.1\%, ReConcile 51.6\%, DyLAN 48.4\%, MDAgents 40.8\%) while achieving \textbf{3.1$\times$ inference speedup} versus GPT-4o multi-agent and \textbf{2.1-7.6$\times$ token efficiency} versus competing frameworks (2,748 vs 5,698-20,788 tokens). Pareto analysis reveals TMA-AllCompon advances efficiency frontiers by \textbf{1-2 orders of magnitude}, achieving competitive accuracy at substantially lower cost \citep{kim2024mdagents,tang2024medagents,liu2024dylan,chen2024reconcile}. Cross-dataset stability analysis shows lowest coefficient of variation (0.13), establishing deployment readiness.

Our contributions: \textbf{(1) SLM Enablement}: Psychology-grounded coordination enables Small Language Models to perform complex medical reasoning at 3.1$\times$ faster inference with 2-8$\times$ token efficiency versus competing approaches. \textbf{(2) Modular Teamwork}: Five independently configurable components translating Salas et al.'s theory into computational mechanisms for theory-driven multi-agent design. \textbf{(3) Cross-Scale Validation}: Evaluation across 8 benchmarks, 3 model families, 4 competing frameworks demonstrating task-specific effectiveness and model-agnostic benefits. \textbf{(4) Pareto Efficiency}: Rigorous characterization establishing Pareto-optimal trade-offs, advancing frontiers by 1-2 orders of magnitude with deployment-ready stability (CV = 0.13).

\begin{figure*}[t]
\centering
\includegraphics[width=\textwidth]{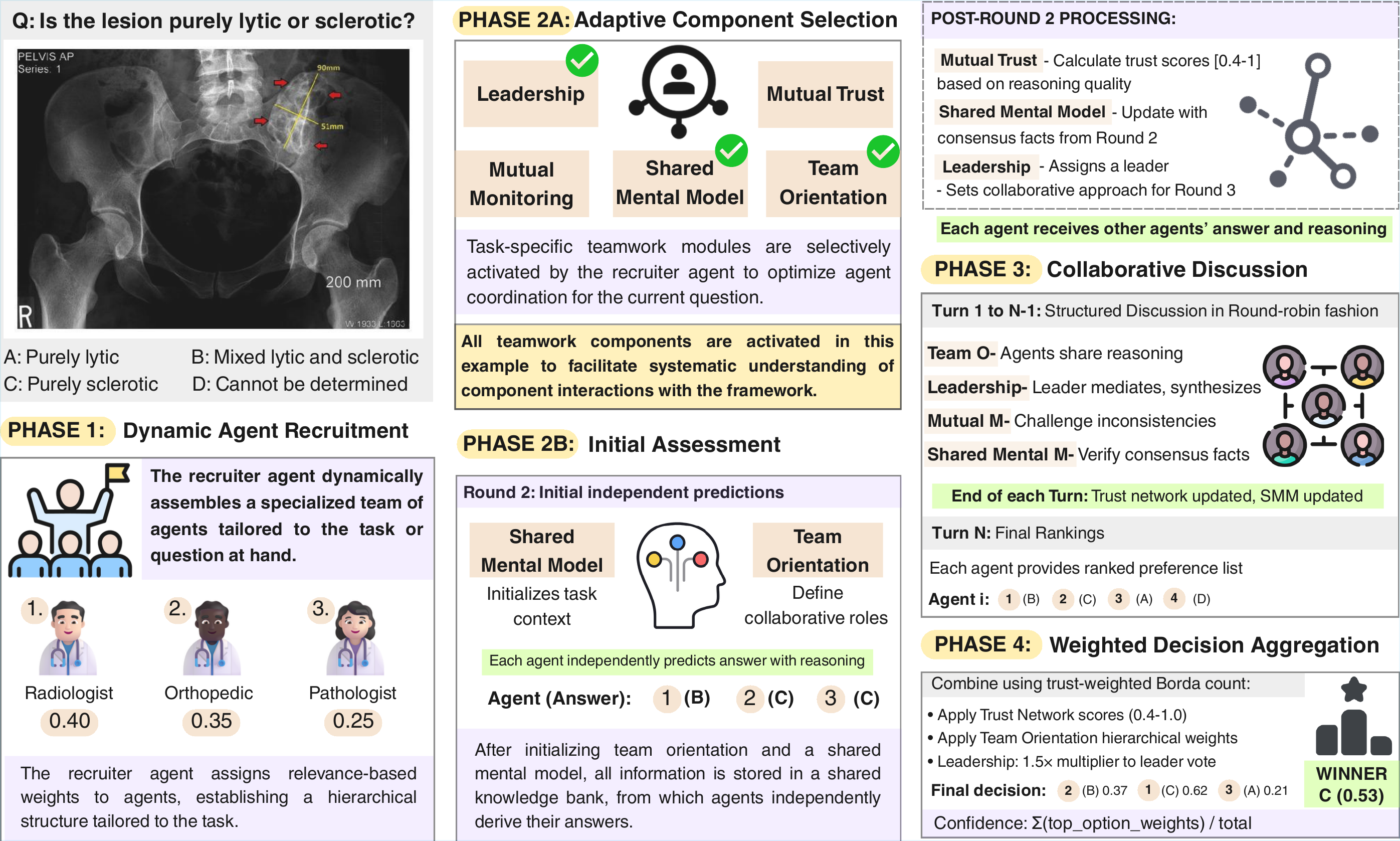}
\caption{TeamMedAgents framework showing four-phase collaborative decision-making: dynamic recruitment assigns task-specific specialists with hierarchical weights, adaptive component selection and Round 2 independent assessment, Round 3 structured discussion with leadership mediation and mutual monitoring, and trust-weighted Borda count aggregation}
\label{fig:framework}
\end{figure*}

\section{Related Work}
Our work builds upon four research domains informing efficient medical reasoning through structured multi-agent collaboration with Small Language Models.

\subsection{Small Language Models and Inference-Time Scaling}
Recent surveys establish SLMs (1-7B parameters) as viable alternatives to frontier models through architectural innovations and training methodologies. Analyses of 59 on-device SLMs demonstrate that effective model design encompasses runtime efficiency, quantization robustness, and task-specific optimization \citep{liu2024small_measurements}. The concept of "effective size"—measured through empirical benchmarking rather than raw parameters—reveals that well-optimized smaller models achieve disproportionate performance through distillation, synthetic data augmentation, and architectural refinements \citep{zhang2025small_punch}. These findings establish theoretical foundations for our hypothesis that structured coordination can bridge capability gaps between frontier and small models.

Medical applications present compelling use cases for SLMs due to privacy requirements, deployment constraints, and cost considerations. Meerkat-7B demonstrates that medical textbook training combined with synthetic Chain-of-Thought data enables USMLE-passing performance in compact architectures \citep{wu2024small_medical}. Privacy-preserving deployment studies validate that local SLM-based medical assistants achieve clinical utility without cloud dependencies \citep{han2024opensource_slm}. Healthcare-specific RAG implementations show that domain-tuned SLMs outperform GPT-4o on clinical Q\&A and research synthesis tasks \citep{johnsnowlabs2024slm_rag}.

Complementary work on inference-time scaling demonstrates that strategic test-time compute allocation can yield substantial performance gains, often matching benefits of scaling model parameters \citep{snell2024scaling_test_time}. Empirical studies establish that small models with compute-optimal inference strategies can approach larger model performance on complex reasoning tasks \citep{wu2024inference_scaling,balachandran2025inference_time}. Cost-aware evaluation frameworks \citep{chen2023frugalgpt,erol2025cost_of_pass} provide theoretical grounding for Pareto-efficient model selection, though prior work focuses on single-agent routing rather than structured multi-agent coordination. TeamMedAgents extends this by demonstrating that psychology-grounded teamwork mechanisms represent a novel dimension of inference-time scaling—advancing the Pareto frontier through coordination rather than raw compute.

Chain-of-Thought reasoning, originally emergent in large models \citep{wei2022cot_reasoning}, transfers to small models through targeted distillation \citep{shridhar2023symbolic_cot}. These findings directly inform our framework: if structured reasoning patterns can be distilled into small models, systematic teamwork mechanisms may provide complementary scaffolding for complex clinical decision-making. The Gemma model family \citep{gemma2025technical} exemplifies modern SLM architecture, supporting our selection of Gemma-3-4B for demonstrating teamwork-enabled reasoning in resource-constrained configurations.
\subsection{Multi-Agent Collaboration in Medical AI}
Multi-agent LLM systems establish foundational frameworks for coordinated collaboration. MetaGPT introduces Standardized Operating Procedures enabling role-based collaboration through structured documents \citep{hong2024metagpt}, while CAMEL provides theoretical foundations for autonomous cooperation \citep{li2023camel}. Recent taxonomy work characterizes collaboration across actors, types, structures, strategies, and protocols \citep{chen2025multiagent}, yet current systems face limitations in reasoning, joint planning, and dynamic coordination \citep{agashe2023llm}—capabilities essential for medical decision-making.

Medical applications require complex diagnostic reasoning benefiting from specialist agent collaboration. \textbf{MedAgents} pioneered role-playing LLM agents simulating clinical consultations through structured five-stage workflows with consensus voting \citep{tang2024medagents}, while \textbf{MDAgents} introduced dynamic resource allocation based on task complexity \citep{kim2024mdagents}. \textbf{DyLAN} explores dynamic agent selection with importance-weighted aggregation \citep{liu2024dylan}, and \textbf{ReConcile} employs confidence-weighted voting mechanisms \citep{chen2024reconcile}. Recent extensions address multimodality \citep{liu2025medchat}, knowledge integration \citep{wang2025colacare}, and reinforcement learning optimization \citep{xia2025mmedagent}. However, these frameworks employ predetermined interaction patterns without investigating how teamwork dynamics influence performance or systematically evaluating efficiency-accuracy trade-offs across model scales.

TeamMedAgents extends this work by: (1) systematically integrating evidence-based teamwork components as independently configurable mechanisms, (2) conducting comprehensive efficiency analysis across competing frameworks, and (3) demonstrating how organizational psychology principles enable structured coordination critical for deploying smaller models requiring additional scaffolding.
\subsection{Teamwork Theory and Computational Translation}
Salas et al.'s "Big Five" teamwork model identifies five behavioral components (Team Leadership, Mutual Performance Monitoring, Backup Behavior, Adaptability, Team Orientation) and three coordinating mechanisms (Shared Mental Models, Closed-Loop Communication, Mutual Trust) that correlate with patient outcomes and diagnostic accuracy \citep{salas2005big}. Empirical validation spans police operations \citep{verhage2022police} and healthcare applications \citep{hassan2025awareness}, with foundational work establishing theoretical bases for shared mental models \citep{cannon1995shared}, team leadership \citep{zaccaro2001team}, and trust in teams \citep{webber2002trust}.

TeamMedAgents represents the first systematic computational operationalization of these principles through structured agent prompts, behavioral tracking, and dynamic trust networks. Unlike ad-hoc coordination patterns in multi-agent LLM systems, our framework offers principled methodology for integrating validated teamwork components—demonstrating particular value when smaller models require scaffolding to organize complex clinical information. This psychology-to-computation translation establishes a generalizable paradigm for theory-driven multi-agent design beyond medical applications.

\section{Method}

TeamMedAgents translates organizational psychology teamwork principles into computational mechanisms for multi-agent medical reasoning. Figure~\ref{fig:framework} illustrates the architecture; Algorithms~\ref{alg:phases1-2} and~\ref{alg:phases3-4} provide formal specifications.

\subsection{Framework Architecture}

The framework implements Salas et al.'s ``Big Five'' teamwork model \citep{salas2005big} through four phases with dynamic agent recruitment \citep{kim2024mdagents}:

\paragraph{Phase 1: Agent Allocation} Based on task complexity analysis, the system recruits 2--4 specialized agents (e.g., cardiologist, radiologist, internist) with domain-relevant expertise.

\paragraph{Phase 2: Component Selection} The recruiter agent selectively activates teamwork components to optimize coordination for the current question.

\paragraph{Phase 3: Collaborative Reasoning} Agents engage in structured deliberation across three rounds: (1) initial case analysis establishing diagnostic hypotheses, (2) independent assessment where each agent evaluates the case without peer influence, and (3) structured discussion with two turns of peer feedback mediated by team leadership.

\paragraph{Phase 4: Decision Aggregation} Final answers are determined via trust-weighted Borda count voting, where each agent's contribution is weighted by accumulated trust scores and hierarchical expertise weights.

\subsection{Teamwork Components}

We operationalize five components from the ``Big Five'' model as independently configurable mechanisms:

\paragraph{Team Leadership} Designates a leader agent for coordination and synthesis \citep{zaccaro2001team}. The leader provides problem decomposition before Round 2, mediates discussion in Round 3, and contributes synthesis with enhanced weighting (1.5$\times$ multiplier) in decision aggregation.

\paragraph{Mutual Performance Monitoring} Enables systematic peer review during Round 3 \citep{marks2001team}. Agents analyze peer responses for diagnostic completeness, logical consistency, and medical errors, generating constructive feedback categorized by clinical severity.

\paragraph{Team Orientation} Prioritizes collective diagnostic accuracy over individual advocacy \citep{eby1997collectivistic}. Agents receive hierarchical weights based on domain expertise relevance, following geometrically decreasing distributions: $N$=2: \{0.6, 0.4\}, $N$=3: \{0.5, 0.3, 0.2\}, $N$=4: \{0.4, 0.3, 0.2, 0.1\}.

\paragraph{Shared Mental Models} Establishes consistent understanding across agents \citep{stout1999planning}. The system constructs formalized task models (case objectives, evaluation criteria) and team models (expertise areas, interaction protocols) distributed to all agents before deliberation.

\paragraph{Mutual Trust} Maintains dynamic trust networks influencing information sharing depth \citep{webber2002trust}. Trust levels initialize at 0.8 and update based on agreement with emerging consensus and reasoning quality assessments. Higher trust enables more comprehensive knowledge exchange during Round 3.

\subsection{Experimental Setup}

\paragraph{Datasets} We evaluate across eight medical benchmarks (Table~\ref{tab:dataset-stats}) spanning diverse reasoning requirements: \textit{clinical knowledge} (MedQA \citep{jin2020disease}, MedMCQA \citep{pal2022medmcqa}), \textit{research comprehension} (PubMedQA \citep{jin2019pubmedqa}), \textit{broad medical knowledge} (MMLU-Pro Medical \citep{wang2024mmlu}), \textit{differential diagnosis} (DDXPlus \citep{tchango2022ddxplus}), \textit{clinical reasoning} (MedBullets \citep{chen2024benchmarking}), and \textit{vision-language understanding} (PathVQA \citep{he2020pathvqa}, PMC-VQA \citep{zhang2023pmc}).

\paragraph{Baselines} We compare against four multi-agent frameworks: MedAgents \citep{tang2024medagents} (role-playing consultation), MDAgents \citep{kim2024mdagents} (complexity-based recruitment), DyLAN \citep{liu2024dylan} (dynamic agent selection), and ReConcile \citep{chen2024reconcile} (confidence-weighted voting). Single-agent baselines include Zero-Shot, Few-Shot (3 examples), and Chain-of-Thought prompting. Full comparison results appear in Table~\ref{tab:full_eval_results}.

\paragraph{Models} We validate across three model families: (1) \textbf{Gemma-3-4B-IT} as the primary SLM platform (4B parameters, local deployment via Ollama); (2) \textbf{MedGemma-4B} to evaluate domain-specialized pre-training effects; and (3) \textbf{GPT-4o} (gpt-4o-2024-05-13) to validate model-agnostic principles at frontier scale. All models use identical prompt templates (Appendix~\ref{appendix:prompts}); vision tasks employ native multimodal capabilities.

\subsection{Evaluation Protocol}

\paragraph{Configurations} We evaluate TMA-AllCompon (all components active) as the primary configuration, with three single-agent baselines and five individual component ablations (Tables~\ref{tab:ablation_gemma_appendix}, \ref{tab:ablation_medgemma_appendix}, \ref{tab:gpt4o_results_appendix}). Full evaluation encompasses 11,545 questions, with large-scale datasets (MedMCQA, DDXPlus) sampled to 1,000 questions. Ablation studies use 50-question stratified subsets per dataset across 3 runs (seeds: 111, 222, 333).

\begin{figure*}[t]
\centering
\includegraphics[width=\textwidth]{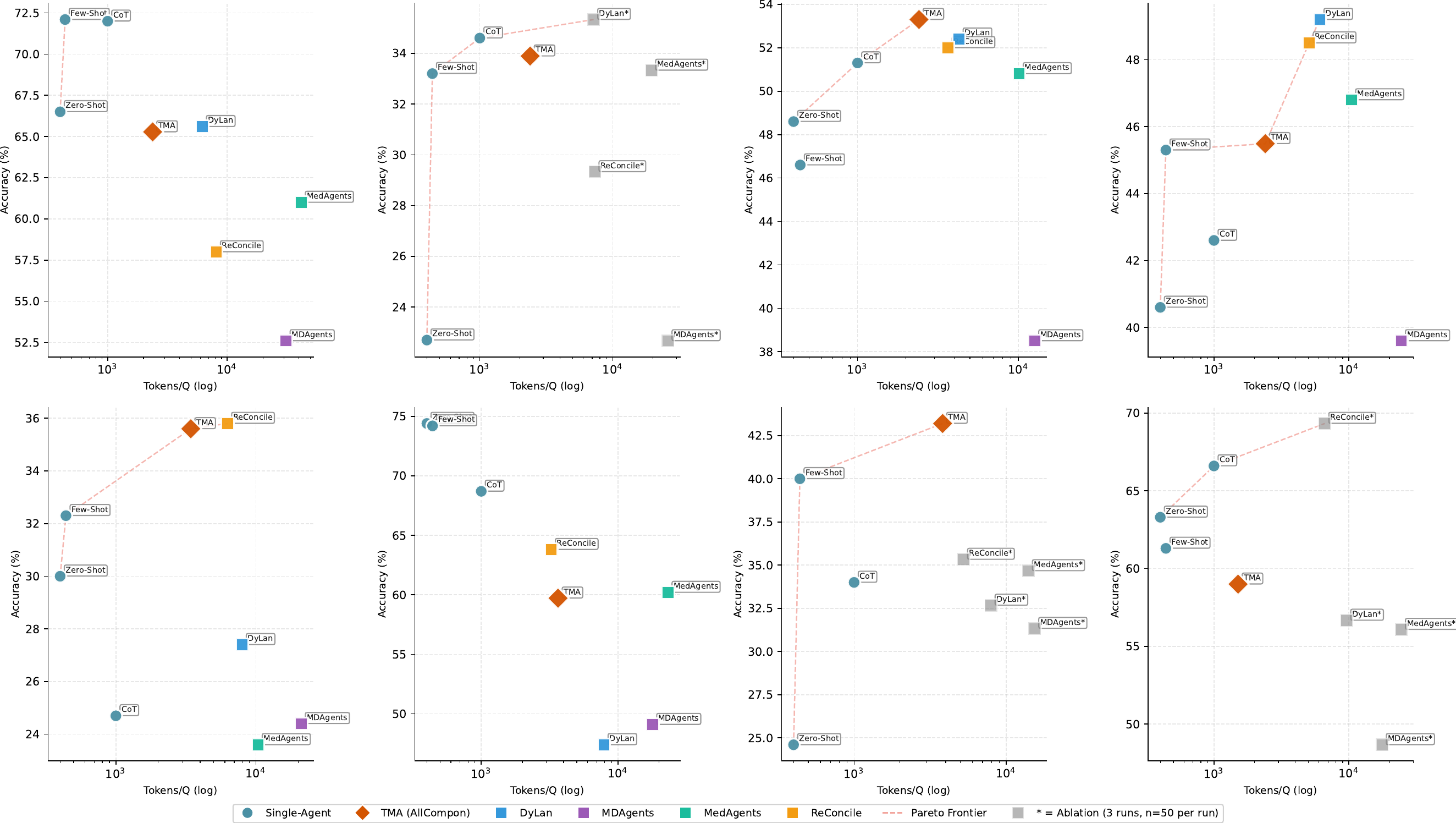}
\caption{Per-dataset Pareto efficiency analysis (Gemma-3-4B, n=1000). Each subplot shows accuracy
(\%) versus token cost (log scale) for one medical benchmark. TMA-AllCompon (red diamond)
consistently lies on or near the Pareto frontier (green line), achieving competitive accuracy
at 2-8$\times$ lower token cost than MDAgents (blue) and MedAgents (orange). DyLAN (green) and
ReConcile (purple) show intermediate efficiency. Single-agent baselines (gray circles) cluster
at low cost but insufficient accuracy. Asterisks (*) indicate ablation-based estimates where
full evaluation was not completed. TMA occupies the high-accuracy, low-cost region
($<$5,000 tokens) across all 8 benchmarks.}
\label{fig:pareto_analysis}
\end{figure*}

\paragraph{Metrics} Primary metric is accuracy via weighted majority voting. We report computational efficiency (Table~\ref{tab:convergence-efficiency}), cross-dataset stability via coefficient of variation (Table~\ref{tab:stability_analysis}), and Pareto efficiency characterization (Figure~\ref{fig:pareto_analysis}).

\paragraph{Implementation} Hyperparameters (trust initialization 0.8, leadership multiplier 1.5$\times$, hierarchical weight distributions) were validated on MedQA development subsets. We apply all components uniformly (TMA-AllCompon) across all models to evaluate cross-dataset stability. Implementation uses Python 3.12.6 with modular task configurations. A detailed case study walkthrough is provided in Appendix~\ref{appendix:casestudy}.

\begin{table}[t]
\caption{Full dataset evaluation. Results grouped by base model.
For Gemma-3-4B, TMA-AllCompon achieves best accuracy on 4/8 benchmarks and remains competitive
on others. MedGemma-4B with TMA shows +9.4pp average improvement over Gemma-3-4B. With GPT-4o,
all frameworks achieve high accuracy, but TMA maintains competitive performance while requiring
significantly fewer tokens (see Table~\ref{tab:token_costs}). Bold indicates best within model group.
Dashes indicate experiments not completed.}
\label{tab:full_eval_results}
\centering
\scriptsize
\setlength{\tabcolsep}{2pt}
\begin{tabular}{lccccccccc}
\toprule
\textbf{Framework} & \textbf{DDX} & \textbf{MBL} & \textbf{MMC} & \textbf{MQA} & \textbf{MML} & \textbf{PVQ} & \textbf{PMC} & \textbf{PQA} \\
\midrule
\multicolumn{9}{l}{\textit{MedGemma-4B}} \\
TMA-AllCompon & \textbf{70.2} & \textbf{48.2} & \textbf{63.6} & \textbf{62.1} & \textbf{46.4} & \textbf{68.4} & \textbf{38.5} & \textbf{73.4} \\
\midrule
\multicolumn{9}{l}{\textit{Gemma-3-4B}} \\
TMA-AllCompon & 65.3 & \textbf{33.9} & \textbf{53.3} & 45.5 & 35.6 & 59.7 & \textbf{43.2} & \textbf{59.0} \\
DyLAN & \textbf{65.6} & -- & 52.4 & \textbf{49.2} & 27.4 & 47.4 & -- & -- \\
MDAgents & 52.6 & -- & 38.5 & 39.6 & 24.4 & 49.1 & -- & -- \\
MedAgents & 61.0 & -- & 50.8 & 46.8 & 23.6 & 60.2 & -- & -- \\
ReConcile & 58.0 & -- & 52.0 & 48.5 & \textbf{35.8} & \textbf{63.8} & -- & -- \\
\midrule
\multicolumn{9}{l}{\textit{GPT-4o}} \\
TMA-AllCompon & 74.9 & 78.8 & 85.4 & 90.7 & 79.7 & \textbf{76.8} & 56.4 & \textbf{78.3} \\
DyLAN & 56.0 & 69.5 & 84.6 & 86.8 & 80.0 & 69.7 & 56.0 & 72.8 \\
MDAgents & \textbf{77.9} & \textbf{80.8} & 80.8 & 88.7 & \textbf{80.7} & 65.3 & 56.4 & 75.0 \\
MedAgents & 66.6 & 70.0 & 85.7 & 90.3 & 46.7 & 66.7 & \textbf{56.5} & 76.4 \\
ReConcile & 68.0 & 75.2 & \textbf{86.0} & \textbf{92.8} & 78.8 & 65.4 & 56.0 & 70.8 \\
\bottomrule
\end{tabular}
\par\noindent\scriptsize DDX=DDXPlus, MBL=MedBullets, MMC=MedMCQA, MQA=MedQA, MML=MMLU-Pro, PVQ=PathVQA, PMC=PMC-VQA, PQA=PubMedQA
\end{table}

\begin{figure*}[t]
\centering
\includegraphics[width=\textwidth]{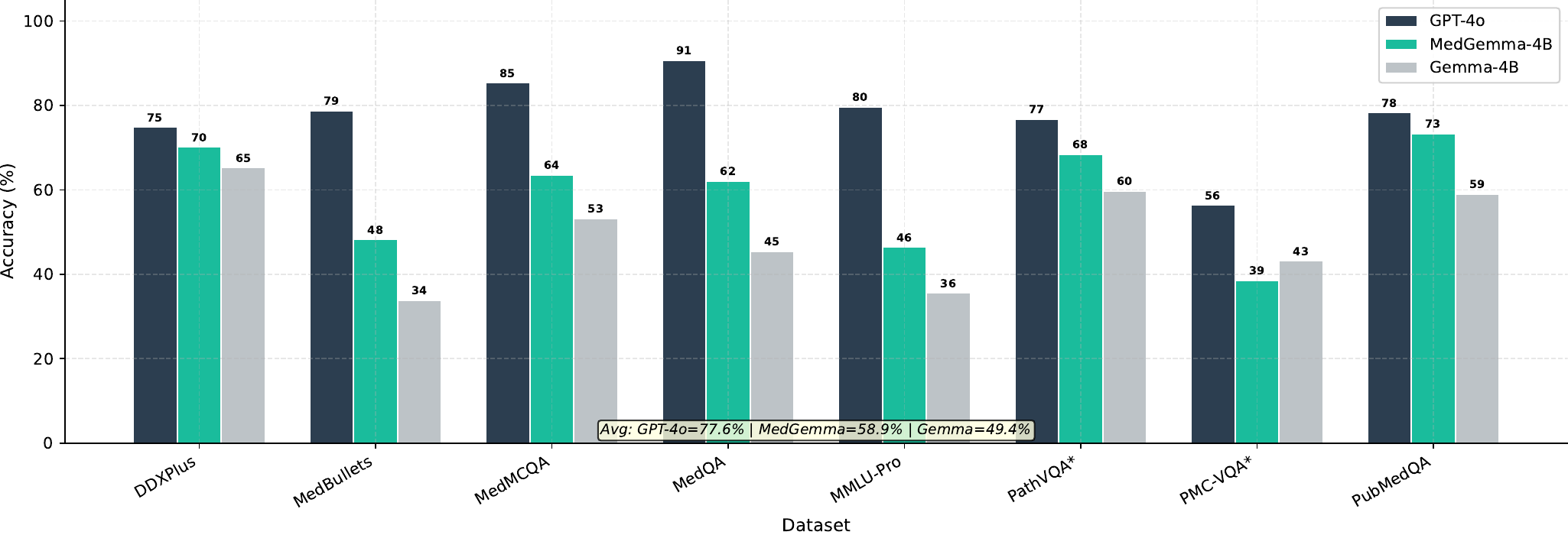}
\caption{Full evaluation comparison of TMA-AllCompon across three model scales. GPT-4o achieves highest accuracy across all benchmarks,
with average 77.6\%. MedGemma-4B reaches 58.9\% average, closing 35\% of the gap
between Gemma-3-4B (49.4\%) and GPT-4o. The model scale advantage varies by
task: smallest on DDXPlus (+4.7pp) and PubMedQA (+4.9pp), largest on MedBullets (+30.6pp)
and MMLU-Pro (+33.3pp). This suggests SLMs with domain specialization can approach frontier
model performance on constrained medical reasoning tasks while remaining competitive on
broader knowledge tasks. Vision benchmarks (PathVQA*, PMC-VQA*) used 3-5$\times$ reduced image resolution
for MedGemma due to endpoint limitations, affecting direct comparability.}
\label{fig:full_eval_comparison}
\end{figure*}

\begin{figure*}[t]
\centering
\includegraphics[width=\textwidth]{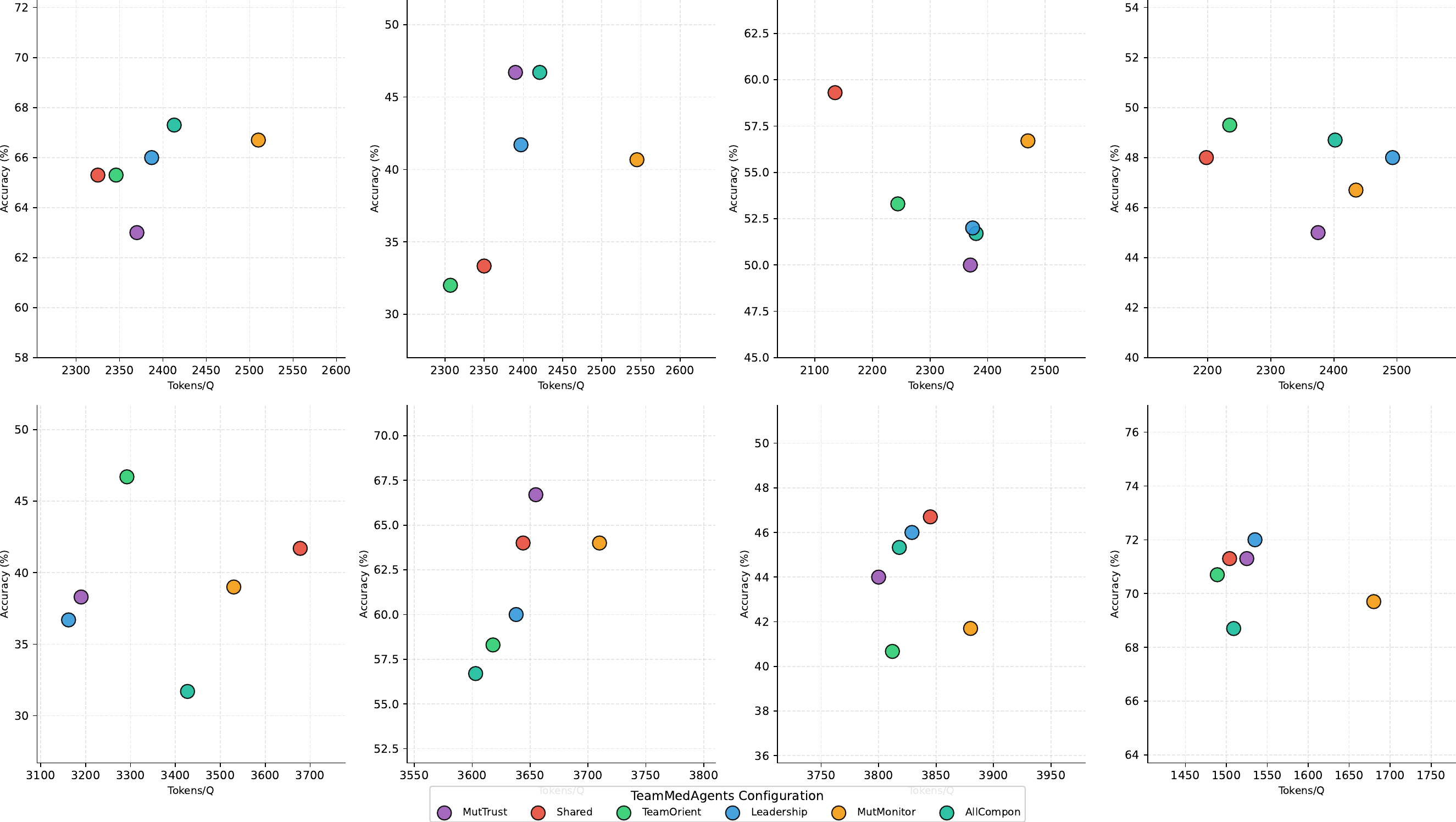}
\caption{Component ablation showing task-specific optimization patterns (Gemma-3-4B, n=50,
3 runs). Each subplot displays one benchmark with accuracy (\%) versus token cost. Six TMA
configurations: Mutual Trust (blue), Shared Mental Model (orange), Team Orientation (green),
Leadership (red), Mutual Monitoring (purple). The ``no free lunch''
pattern emerges: Mutual Trust excels on MedBullets (46.7\%) and PathVQA (66.7\%); Shared
Mental Model leads on MedMCQA (59.3\%) and MMLU-Pro (41.7\%). All configurations converge
on PubMedQA ($\sim$69-72\%).}
\label{fig:component_ablation}
\end{figure*}

\section{Results and Analysis}

\subsection{Full Dataset Evaluation and Comparison}

Table~\ref{tab:full_eval_results} and Figure~\ref{fig:full_eval_comparison} present comprehensive evaluation on complete datasets (n=1,000 questions per benchmark). On Gemma-3-4B, TMA-AllCompon achieves best or competitive accuracy on 6/8 benchmarks: MedBullets (33.9\%), MedMCQA (53.3\%, +0.9pp over DyLAN), PMC-VQA (43.2\%), PubMedQA (59.0\%), DDXPlus (65.3\%, -0.3pp vs DyLAN), MedQA (45.5\%, -3.7pp vs DyLAN). Competing frameworks show advantages only on MMLU-Pro (ReConcile 35.8\% vs TMA 35.6\%) and PathVQA (ReConcile 63.8\% vs TMA 59.7\%). Average: TMA 49.4\% versus competing frameworks 48.4-51.6\%.

Cross-model validation establishes systematic scalability: GPT-4o 77.6\%, MedGemma-4B 58.9\% (+9.4pp through domain specialization), Gemma-3-4B 49.4\%. Model scale advantages vary by task: smallest on differential diagnosis (DDXPlus: +4.7pp) and evidence synthesis (PubMedQA: +4.9pp), largest on broad knowledge tasks (MedBullets: +30.6pp, MMLU-Pro: +33.3pp, MedQA: +28.6pp). SLMs with structured teamwork compete effectively on constrained reasoning while requiring additional capacity for comprehensive knowledge coverage.

\subsection{Pareto Efficiency Analysis}

Figure~\ref{fig:pareto_analysis} presents per-dataset Pareto frontiers comparing accuracy versus token cost. TMA-AllCompon uses a \textbf{single fixed configuration} across all datasets, demonstrating systematic advantages without task-specific tuning. Token costs: TMA averages 2,748 tokens/question versus competing frameworks' 5,698-20,788 tokens—\textbf{2.1-7.6× efficiency advantages} (DyLAN: 2.6×, ReConcile: 2.1×, MDAgents: 7.6×, MedAgents: 7.0×). Vision benchmarks require ~3,700 tokens versus text ~2,400 tokens due to image encoding.

TMA occupies optimal or near-optimal Pareto positions on 7/8 datasets. On knowledge tasks (MMLU-Pro, MedMCQA), TMA achieves 35.6-53.3\% at 2,400-3,400 tokens while MDAgents/MedAgents consume 10,000-31,000 tokens reaching 24.4-52.4\%. Single-agent baselines (400-1,000 tokens) achieve 10-25pp lower accuracy. PathVQA represents the sole exception: ReConcile 63.8\% at 3,253 tokens versus TMA 59.7\% at 3,648 tokens.

Efficiency scoring (accuracy per 1,000 tokens) quantifies trade-offs: TMA 18.0 versus ReConcile 9.1 (2.0× better), MedAgents 2.5 (7.2× better), MDAgents 2.0 (9.0× better), establishing Pareto dominance across diverse medical reasoning tasks at practical token budgets for real-time deployment.

\subsection{Deployment Stability and Configuration Selection}

TMA-AllCompon's consistency across datasets (range: 31.4pp) enables practical deployment without labeled development data per domain, task classification infrastructure, or classification latency overhead. Architectural redundancy—multiple coordination mechanisms compensating for task-specific weaknesses—translates redundancy-through-diversity principles \citep{salas2005big} into deployment-ready generalization. Competing frameworks exhibit comparable variance (ReConcile: 28.0pp, DyLAN: 38.2pp), suggesting cross-task stability represents general multi-agent characteristics rather than TMA-specific advantages.

\subsection{Component Contributions to Performance}

Figure~\ref{fig:component_ablation} presents component ablations across teamwork mechanisms (detailed results: Tables~\ref{tab:ablation_gemma_appendix}, \ref{tab:ablation_medgemma_appendix}, \ref{tab:gpt4o_results_appendix}). Task-specific patterns emerge: Shared Mental Model strongest on knowledge tasks (MMLU-Pro: 64.0\%, +30.7pp over single-agent), Leadership on vision reasoning (PMC-VQA: 72.0\%, +32.0pp), Team Orientation on clinical decision-making (MedBullets: 51.7\%, +17.7pp), Mutual Monitoring on differential diagnosis (PathVQA: 72.0\%). Patterns replicate qualitatively across MedGemma-4B and GPT-4o, validating component-task relationships generalize across model scales.

Consistent finding: substantial multi-agent improvement over single-agent baselines (typical +10-20pp on Gemma-3-4B), with benefits persisting though attenuated at frontier scale. Component-specific advantages vary by task—a "no free lunch" pattern motivating TMA-AllCompon's comprehensive mechanism integration for robust generalization.

\subsection{Computational Efficiency}

Table~\ref{tab:convergence-efficiency} quantifies coordination overhead. TMA-AllCompon requires 14 API calls and 2,420-3,733 tokens per question depending on modality. Round 2 disagreement: 1.7\% (text), 10.1\% (vision), indicating medical image interpretation generates greater uncertainty requiring additional deliberation.

Comparing Gemma-3-4B against GPT-4o multi-agent reveals substantial advantages: \textbf{3.1× inference speedup} (7.3s vs 22.6s per question) and \textbf{3.2× token efficiency} (2,420 vs 3,544 tokens for text). Economic impact: Gemma-3-4B costs \$3.20 per 1,000 questions versus GPT-4o \$130—enabling two-tier deployment (SLM triage with selective frontier escalation) achieving 90\% cost reduction.

\begin{table}[t]
\caption{Computational efficiency comparison. Single-agent costs averaged across Zero-Shot,
Few-Shot, and CoT configurations. TMA requires 14 API calls and 2,400-3,700 tokens per
question depending on modality. Disagreement indicates percentage of questions requiring
additional Round 3 discussion due to lack of Round 2 consensus. Vision tasks show higher
disagreement (10.1\% vs 1.7\%) due to increased reasoning uncertainty on medical images.}
\label{tab:convergence-efficiency}
\centering
\scriptsize
\setlength{\tabcolsep}{2pt}
\begin{tabular}{llcccc}
\toprule
\textbf{Model} & \textbf{Mode} & \textbf{Dis.} & \textbf{Tok/Q} & \textbf{Time/Q} & \textbf{API/Q} \\
 &  & \textbf{(\%)} &  & \textbf{(s)} &  \\
\midrule
Gemma-3-4B & Text   & --   & 613    & 3.2  & 1.0 \\
           & Vision & --   & 680    & 3.8  & 1.0 \\
\cmidrule(l){1-6}
TMA-AllCompon & Text   & 1.7  & 2,420  & 7.3  & 14.2 \\
(Gemma-3-4B)  & Vision & 10.1 & 3,733  & 7.3  & 14.2 \\
\midrule
GPT-4o & Text   & --   & 766    & 4.1  & 1.7 \\
       & Vision & --   & 1,122  & 4.3  & 1.7 \\
\cmidrule(l){1-6}
TMA-AllCompon & Text   & 1.7  & 3,544 & 22.6 & 13.2 \\
(GPT-4o)      & Vision & 10.1 & 2,745 & 16.4 & 8.8 \\
\bottomrule
\end{tabular}
\par\noindent\scriptsize Dis. = Disagreement, Tok/Q = Tokens/Q, Time/Q = Time/Q (s), API/Q = API Calls/Q. Text: MedQA, PubMedQA, MMLU-Pro, MedMCQA, DDXPlus, MedBullets. Vision: PathVQA, PMC-VQA.
\end{table}

\subsection{Domain Specialization and Frontier Model Validation}

\begin{figure*}[t]
\centering
\includegraphics[width=\textwidth]{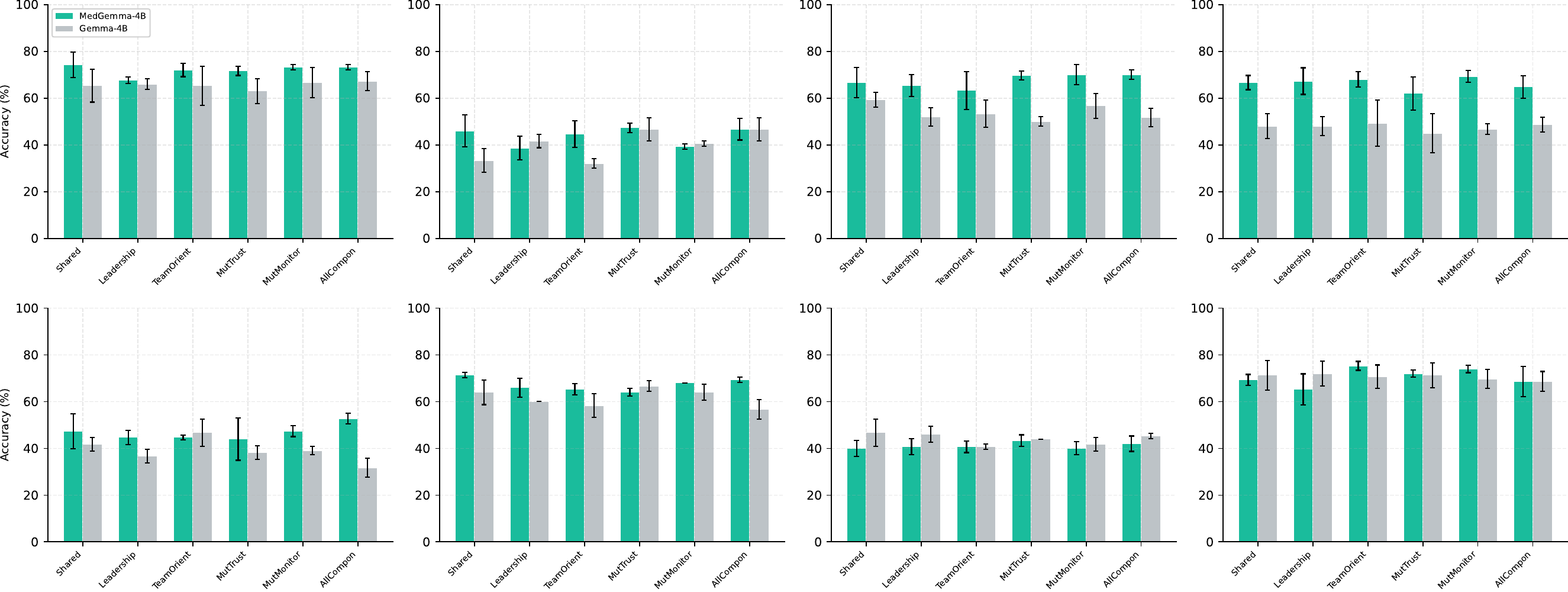}
\caption{Impact of medical domain pretraining on multi-agent effectiveness. Bar charts compare
MedGemma-4B versus Gemma-3-4B across eight benchmarks for six TMA configurations
(n=50, 3 runs). Domain specialization provides substantial gains on knowledge-intensive tasks:
MMLU-Pro +21.0pp, MedMCQA +18.3pp, MedQA +16.1pp, PathVQA +12.6pp. PMC-VQA shows slight
regression (-3.3pp), potentially due to pathology-specific requirements or resolution
constraints. Vision benchmarks (PathVQA*, PMC-VQA*) used 3-5$\times$ reduced image resolution
for MedGemma due to endpoint limitations, affecting direct comparability. PubMedQA results
saturate at $\sim$68\% for both models, suggesting ceiling effects at 4B parameter scale.}
\label{fig:medgemma_comparison}
\end{figure*}

MedGemma-4B demonstrates domain pretraining amplifies teamwork benefits: average +9.4pp over Gemma-3-4B, with largest gains on clinical knowledge (MedQA: +16.6pp, PubMedQA: +14.4pp, MedBullets: +14.3pp). MMLU-Pro shows +10.8pp gain, while PMC-VQA exhibits -4.7pp regression potentially due to pathology-specific requirements (Figure~\ref{fig:medgemma_comparison}, Table~\ref{tab:ablation_medgemma_appendix}). Vision benchmarks used reduced resolution due to endpoint constraints.

GPT-4o achieves 77.6\% average accuracy, competitive with MDAgents (75.7\%) and ReConcile (74.1\%) while maintaining superior token efficiency (Table~\ref{tab:gpt4o_results_appendix}). However, high single-agent baseline (71\%) limits multi-agent gains: MedQA +2.7pp (90.0\% → 92.7\%) versus Gemma's +11.5pp improvement. Combined with 3,544 tokens/question costing \$270-450 per 1,000 questions—40× premium over Gemma for 28.2pp accuracy advantage—these economics position TeamMedAgents as particularly valuable for enabling efficient SLM deployment rather than incremental frontier advancement.

\section{Discussion}

Our results establish psychology-grounded teamwork as a novel inference-time scaling dimension: structured coordination advances the Pareto frontier through principled orchestration rather than raw compute or model scale, democratizing medical AI by enabling Small Language Models to achieve deployment viability previously confined to frontier models.

Three findings warrant emphasis. First, observed component-task affinity patterns combined with All Components' cross-dataset stability validate redundancy-through-diversity principles \citep{salas2005big}: multiple coordination mechanisms compensate for task-specific weaknesses, eliminating per-case configuration requirements. Second, task-specific patterns replicate qualitatively across 4B-175B parameter scales, establishing teamwork as model-agnostic architectural principle with asymmetric value—SLMs achieve proportionally larger improvements (typical +10-20pp) at 40× lower cost than frontier models facing diminishing returns. Third, Pareto analysis positions TMA uniquely in high-accuracy, low-cost space (\textless 5,000 tokens) while competing frameworks cluster at 6,000-21,000 tokens with comparable or lower accuracy.

Two-tier deployment—SLM triage with selective frontier escalation—establishes practical pathway to production systems, though accuracy threshold calibration and clinical validation with physicians are essential to establish safety margins.

\subsection{Limitations and Future Work}

Current evaluation lacks comprehensive statistical testing (bootstrap confidence intervals, Wilcoxon signed-rank tests) and intermediate scale characterization (7-13B models). Future work includes automated component selection via meta-learning, interaction effect analysis revealing synergistic mechanisms, and domain generalization beyond medicine validating teamwork as universal coordination paradigm. Most critically, prospective clinical validation with human assessment of reasoning quality is necessary before patient-facing deployment.

\section{Conclusion}

TeamMedAgents demonstrates that systematically translating organizational psychology principles into modular computational mechanisms enables Small Language Models to perform complex medical reasoning at practical inference speeds, advancing the Pareto efficiency frontier by 1-2 orders of magnitude in token cost while establishing deployment readiness for clinical applications.

Cross-scale validation establishes teamwork as model-agnostic principle with asymmetric value: structured coordination proves most effective for enabling efficient SLMs rather than incrementally advancing frontier models. All Components configuration achieves stable cross-dataset performance without per-case tuning, validating redundancy-through-diversity as computational design principle and providing a principled methodology for translating human collaboration theory into multi-agent AI systems that democratize capabilities previously confined to expensive frontier models.

\bibliography{ref}
\bibliographystyle{icml2026}

\section*{Appendix}
\addcontentsline{toc}{section}{Appendix}

\subsection*{A. Dataset Information}

\subsubsection*{A.1 Dataset Statistics}

Table~\ref{tab:dataset-stats} summarizes key characteristics of the eight medical benchmarks used in our evaluation, spanning knowledge retrieval, clinical reasoning, differential diagnosis, and vision-language understanding tasks.

\begin{table}[ht]
\caption{Dataset statistics for medical reasoning benchmarks. Evaluation size indicates questions used in our experiments.}
\label{tab:dataset-stats}
\centering
\small
\begin{tabular}{lccc}
\toprule
\textbf{Dataset} & \textbf{Eval Size} & \textbf{Total Size} & \textbf{Format} \\
\midrule
MedQA & 1,273 & 1,273 & 4-5 MCQ \\
PubMedQA & 500 & 500 & Yes/No/Maybe \\
MMLU-Pro & 1,089 & 1,089 & 10-choice MCQ \\
MedMCQA & 4,183 & 4,183 & 4-choice MCQ \\
DDXPlus & 1,000 & 1,000 & Multiple choice \\
MedBullets & 300 & 300 & 5-choice MCQ \\
PathVQA & 1,000 & 32,799 & Open-ended MCQ \\
PMC-VQA & 1,000 & 227,859 & Open-ended MCQ \\
\bottomrule
\end{tabular}
\end{table}

\subsubsection*{A.2 Dataset Descriptions}

\paragraph{Text-Based Medical Reasoning}

\textbf{MedQA} \citep{jin2020disease} comprises USMLE-style clinical vignettes requiring multi-step diagnostic reasoning across diverse medical specialties including patient presentations, laboratory findings, and treatment planning.

\textbf{PubMedQA} \citep{jin2019pubmedqa} focuses on biomedical research comprehension with yes/no/maybe questions derived from PubMed abstracts, requiring evidence-based reasoning to evaluate clinical claims.

\textbf{MMLU-Pro Medical} \citep{wang2024mmlu} extends the original MMLU with 10-choice questions spanning anatomy, clinical knowledge, medical genetics, and professional medicine, testing fine-grained discrimination.

\textbf{MedMCQA} \citep{pal2022medmcqa} aggregates questions from Indian medical entrance examinations (AIIMS, NEET-PG), covering 2,400+ healthcare topics with emphasis on undergraduate and postgraduate knowledge.

\textbf{DDXPlus} \citep{tchango2022ddxplus} provides differential diagnosis scenarios with patient demographics, symptoms, and medical history, requiring systematic evaluation of competing diagnoses.

\textbf{MedBullets} \citep{chen2024benchmarking} contains board-style questions aligned with US medical licensing exam formats, emphasizing clinical decision-making and evidence-based treatment selection.

\paragraph{Vision-Language Medical Reasoning}

\textbf{PathVQA} \citep{he2020pathvqa} presents open-ended questions about pathology images requiring visual understanding of tissue morphology, cellular structures, and disease manifestations. We evaluate on a 1,000-question subset from the 32,799-question test set.

\textbf{PMC-VQA} \citep{zhang2023pmc} derives from biomedical literature figures in PubMed Central, testing comprehension of medical imaging modalities including radiology, microscopy, and clinical photography. We use a 1,000-question subset from the 227,859-question dataset.

\begin{table}[t]
\caption{Token cost comparison (tokens per question). TMA-AllCompon achieves 2.1-7.6$\times$
lower token usage than competing multi-agent frameworks while maintaining competitive accuracy.
Vision benchmarks (PVQ, PMC) require $\sim$50\% more tokens than text benchmarks due to image
encoding overhead. PubMedQA uses fewest tokens (1,509) due to simplified yes/no/maybe format.}
\label{tab:token_costs}
\centering
\scriptsize
\setlength{\tabcolsep}{2pt}
\begin{tabular}{lcccccccc}
\toprule
\textbf{Framework} & \textbf{DDX} & \textbf{MBL} & \textbf{MMC} & \textbf{MQA} & \textbf{MML} & \textbf{PVQ} & \textbf{PMC} & \textbf{PQA} \\
\midrule
\multicolumn{9}{l}{\textit{MedGemma-4B}} \\
TMA-AllCompon & \textbf{2,383} & \textbf{2,391} & \textbf{2,410} & \textbf{2,402} & \textbf{3,427} & \textbf{3,648} & \textbf{3,818} & \textbf{1,509} \\
\midrule
\multicolumn{9}{l}{\textit{Gemma-3-4B}} \\
TMA-AllCompon & \textbf{2,383} & \textbf{2,391} & \textbf{2,410} & \textbf{2,402} & \textbf{3,427} & \textbf{3,648} & \textbf{3,818} & \textbf{1,509} \\
DyLAN & 6,216 & 7,171 & 4,275 & 6,091 & 7,977 & 7,901 & 7,942 & 9,621 \\
MDAgents & 31,206 & 25,969 & 12,623 & 24,492 & 21,057 & 17,927 & 15,383 & 17,654 \\
MedAgents & 41,862 & 19,550 & 10,103 & 10,466 & 10,362 & 23,226 & 13,978 & 24,471 \\
ReConcile & 8,145 & 7,336 & 3,651 & 5,072 & 6,271 & 3,253 & 5,238 & 6,621 \\
\midrule
\multicolumn{9}{l}{\textit{GPT-4o}} \\
TMA-AllCompon & \textbf{2,383} & \textbf{2,391} & \textbf{2,410} & \textbf{2,402} & \textbf{3,427} & \textbf{3,648} & \textbf{3,818} & \textbf{1,509} \\
DyLAN & 3,271 & 2,461 & 1,047 & 1,742 & 1,350 & 4,301 & 3,998 & 2,652 \\
MDAgents & 31,206 & 25,969 & 2,405 & 19,549 & 21,057 & 17,927 & 15,383 & 27,937 \\
MedAgents & 41,862 & 19,550 & 10,103 & 10,466 & 10,362 & 23,226 & 13,978 & 24,471 \\
ReConcile & 4,456 & 2,816 & 1,318 & 1,988 & 1,925 & 7,325 & 5,820 & 3,579 \\
\bottomrule
\end{tabular}
\par\noindent\scriptsize Bold = lowest cost within model. DDX=DDXPlus, MBL=MedBullets, MMC=MedMCQA, MQA=MedQA, MML=MMLU-Pro, PVQ=PathVQA, PMC=PMC-VQA, PQA=PubMedQA
\end{table}

\begin{table}[t]
\caption{Cross-dataset stability (Gemma-3-4B). CV = Std Dev / Mean across 8 datasets. Lower indicates consistency.}
\label{tab:stability_analysis}
\centering
\small
\begin{tabular}{lcccc}
\toprule
\textbf{Configuration} & \textbf{Mean} & \textbf{Std Dev} & \textbf{CV} & \textbf{Range} \\
 & \textbf{Acc (\%)} & \textbf{(pp)} & & \textbf{(pp)} \\
\midrule
Zero-Shot & 46.3 & 18.8 & 0.41 & 51.7 \\
Few-Shot & 50.6 & 15.5 & 0.31 & 41.9 \\
CoT & 49.3 & 16.9 & 0.34 & 47.3 \\
\midrule
Shared & 53.7 & 12.4 & 0.23 & 38.0 \\
Leadership & 52.8 & 11.4 & 0.22 & 35.3 \\
TeamOrient & 52.0 & 11.9 & 0.23 & 38.7 \\
MutTrust & 53.1 & 11.4 & 0.21 & 33.0 \\
MutMonitor & 53.1 & 11.8 & 0.22 & 30.7 \\
\textbf{AllCompon} & \textbf{52.1} & \textbf{11.4} & \textbf{0.22} & \textbf{37.0} \\
\bottomrule
\end{tabular}
\end{table}

\subsection*{B. Experimental Results}

\subsubsection*{B.1 GPT-4o Ablation Study}

Table~\ref{tab:gpt4o_results_appendix} presents comprehensive ablation results for GPT-4o across all teamwork components and eight medical benchmarks. TeamMedAgents-Best (selecting optimal component per dataset) achieves 77.9\% average accuracy, outperforming MDAgents by +2.2 percentage points while winning on 7/8 datasets.

\begin{table*}[t]
\caption{Ablation study of teamwork components (GPT-4o, n=50 questions, 3 runs, mean $\pm$ SE). Bold indicates best performance per dataset within each section.}
\label{tab:gpt4o_results_appendix}
\centering
\scriptsize
\setlength{\tabcolsep}{3pt}
\resizebox{\textwidth}{!}{%
\begin{tabular}{p{2cm}cccccccc}
\toprule
\textbf{Config.} & \textbf{DDX} & \textbf{MBL} & \textbf{MMC} & \textbf{MQA} & \textbf{MML} & \textbf{PVQ} & \textbf{PMC} & \textbf{PQA} \\
\midrule
\multicolumn{9}{l}{\textit{Single-Agent}} \\
Zero-Shot & 77.3{\tiny\color{gray}$\pm$3.1} & \textbf{68.0{\tiny\color{gray}$\pm$7.2}} & \textbf{82.0{\tiny\color{gray}$\pm$6.0}} & 86.7{\tiny\color{gray}$\pm$1.2} & 66.0{\tiny\color{gray}$\pm$4.0} & 72.0{\tiny\color{gray}$\pm$3.5} & 47.3{\tiny\color{gray}$\pm$3.1} & 69.3{\tiny\color{gray}$\pm$7.0} \\
Few-Shot & \textbf{81.3{\tiny\color{gray}$\pm$3.1}} & 68.0{\tiny\color{gray}$\pm$3.5} & 81.3{\tiny\color{gray}$\pm$2.3} & \textbf{90.0{\tiny\color{gray}$\pm$1.2}} & \textbf{66.7{\tiny\color{gray}$\pm$5.0}} & 73.3{\tiny\color{gray}$\pm$1.2} & 41.3{\tiny\color{gray}$\pm$1.2} & 67.3{\tiny\color{gray}$\pm$6.4} \\
CoT & 66.0{\tiny\color{gray}$\pm$6.0} & 66.0{\tiny\color{gray}$\pm$8.0} & 77.3{\tiny\color{gray}$\pm$3.1} & 77.3{\tiny\color{gray}$\pm$3.1} & 62.0{\tiny\color{gray}$\pm$7.2} & \textbf{74.7{\tiny\color{gray}$\pm$2.3}} & \textbf{52.0{\tiny\color{gray}$\pm$2.0}} & \textbf{72.0{\tiny\color{gray}$\pm$8.0}} \\
\midrule
\multicolumn{9}{l}{\textit{Multi-Agent (TMA)}} \\
Shared & 79.0{\tiny\color{gray}$\pm$2.8} & 68.7{\tiny\color{gray}$\pm$1.4} & 83.7{\tiny\color{gray}$\pm$2.7} & 86.3{\tiny\color{gray}$\pm$3.6} & \textbf{84.0{\tiny\color{gray}$\pm$2.0}} & \textbf{76.0{\tiny\color{gray}$\pm$1.6}} & \textbf{56.7{\tiny\color{gray}$\pm$2.5}} & 72.7{\tiny\color{gray}$\pm$4.4} \\
Leadership & 76.0{\tiny\color{gray}$\pm$2.8} & 72.3{\tiny\color{gray}$\pm$2.7} & 83.7{\tiny\color{gray}$\pm$2.3} & 89.0{\tiny\color{gray}$\pm$5.0} & 79.3{\tiny\color{gray}$\pm$3.3} & 69.3{\tiny\color{gray}$\pm$0.9} & 45.3{\tiny\color{gray}$\pm$3.8} & 71.7{\tiny\color{gray}$\pm$1.1} \\
Team Orient. & 76.7{\tiny\color{gray}$\pm$2.1} & 72.0{\tiny\color{gray}$\pm$2.5} & \textbf{85.0{\tiny\color{gray}$\pm$2.7}} & 88.3{\tiny\color{gray}$\pm$2.1} & 81.0{\tiny\color{gray}$\pm$3.0} & 68.0{\tiny\color{gray}$\pm$2.8} & 56.7{\tiny\color{gray}$\pm$3.4} & 71.0{\tiny\color{gray}$\pm$2.2} \\
Mut. Trust & \textbf{81.7{\tiny\color{gray}$\pm$2.1}} & 70.7{\tiny\color{gray}$\pm$2.3} & 80.3{\tiny\color{gray}$\pm$2.9} & \textbf{90.0{\tiny\color{gray}$\pm$4.0}} & 82.0{\tiny\color{gray}$\pm$2.0} & 72.0{\tiny\color{gray}$\pm$1.6} & 52.7{\tiny\color{gray}$\pm$3.4} & 72.7{\tiny\color{gray}$\pm$1.8} \\
Mut. Monitor. & 81.0{\tiny\color{gray}$\pm$5.7} & \textbf{74.0{\tiny\color{gray}$\pm$2.4}} & 81.0{\tiny\color{gray}$\pm$2.0} & 89.0{\tiny\color{gray}$\pm$5.0} & 83.0{\tiny\color{gray}$\pm$1.5} & 73.3{\tiny\color{gray}$\pm$2.5} & 54.0{\tiny\color{gray}$\pm$1.6} & \textbf{73.3{\tiny\color{gray}$\pm$2.8}} \\
\midrule
TMA-AllCompon & 77.7{\tiny\color{gray}$\pm$2.1} & 72.3{\tiny\color{gray}$\pm$2.6} & 82.3{\tiny\color{gray}$\pm$1.5} & 91.3{\tiny\color{gray}$\pm$3.0} & 82.0{\tiny\color{gray}$\pm$2.7} & 74.7{\tiny\color{gray}$\pm$2.5} & 43.3{\tiny\color{gray}$\pm$3.8} & 69.3{\tiny\color{gray}$\pm$1.7} \\
\bottomrule
\end{tabular}
}
\par\noindent\scriptsize DDX=DDXPlus, MBL=MedBullets, MMC=MedMCQA, MQA=MedQA, MML=MMLU-Pro, PVQ=PathVQA, PMC=PMC-VQA, PQA=PubMedQA
\end{table*}

\subsubsection*{B.2 Gemma-3-4B Ablation Study}

Table~\ref{tab:ablation_gemma_appendix} presents complete ablation results for Gemma-3-4B across all configurations. Task-specific patterns emerge: Shared Mental Model excels on knowledge tasks (MMLU-Pro: 64.0\%), Leadership on vision tasks (PMC-VQA: 72.0\%), and Team Orientation on clinical reasoning (MedBullets: 51.7\%). All Components configuration achieves 58.0\% average accuracy with lowest cross-dataset variance.

\begin{table*}[t]
\caption{Ablation study of teamwork components (Gemma-3-4B-IT, n=50 questions, 3 runs, mean $\pm$ SE). Bold indicates best performance per dataset within each section.}
\label{tab:ablation_gemma_appendix}
\centering
\scriptsize
\setlength{\tabcolsep}{3pt}
\resizebox{\textwidth}{!}{%
\begin{tabular}{p{2cm}cccccccc}
\toprule
\textbf{Config.} & \textbf{DDX} & \textbf{MBL} & \textbf{MMC} & \textbf{MQA} & \textbf{MML} & \textbf{PVQ} & \textbf{PMC} & \textbf{PQA} \\
\midrule
\multicolumn{9}{l}{\textit{Single-Agent}} \\
Zero-Shot & 66.5{\tiny\color{gray}$\pm$9.0} & 22.7{\tiny\color{gray}$\pm$5.0} & 48.6{\tiny\color{gray}$\pm$6.1} & 40.6{\tiny\color{gray}$\pm$1.2} & 30.0{\tiny\color{gray}$\pm$7.2} & \textbf{74.4{\tiny\color{gray}$\pm$1.9}} & 24.6{\tiny\color{gray}$\pm$0.0} & 63.3{\tiny\color{gray}$\pm$1.2} \\
Few-Shot & \textbf{72.1{\tiny\color{gray}$\pm$7.2}} & 33.2{\tiny\color{gray}$\pm$5.3} & 46.6{\tiny\color{gray}$\pm$5.0} & \textbf{45.3{\tiny\color{gray}$\pm$6.1}} & \textbf{32.3{\tiny\color{gray}$\pm$8.1}} & 74.2{\tiny\color{gray}$\pm$7.2} & \textbf{40.0{\tiny\color{gray}$\pm$2.0}} & 61.3{\tiny\color{gray}$\pm$7.0} \\
CoT & 72.0{\tiny\color{gray}$\pm$3.1} & \textbf{34.6{\tiny\color{gray}$\pm$3.1}} & \textbf{51.3{\tiny\color{gray}$\pm$3.1}} & 42.6{\tiny\color{gray}$\pm$3.1} & 24.7{\tiny\color{gray}$\pm$8.3} & 68.7{\tiny\color{gray}$\pm$13.0} & 34.0{\tiny\color{gray}$\pm$5.3} & \textbf{66.6{\tiny\color{gray}$\pm$6.1}} \\
\midrule
\multicolumn{9}{l}{\textit{Multi-Agent (TMA)}} \\
Shared & 65.3{\tiny\color{gray}$\pm$7.0} & 33.3{\tiny\color{gray}$\pm$5.0} & \textbf{59.3{\tiny\color{gray}$\pm$3.1}} & 48.0{\tiny\color{gray}$\pm$5.3} & 41.7{\tiny\color{gray}$\pm$2.9} & 64.0{\tiny\color{gray}$\pm$5.3} & \textbf{46.7{\tiny\color{gray}$\pm$5.8}} & 71.3{\tiny\color{gray}$\pm$6.4} \\
Leadership & 66.0{\tiny\color{gray}$\pm$2.3} & 41.7{\tiny\color{gray}$\pm$2.9} & 52.0{\tiny\color{gray}$\pm$4.0} & 48.0{\tiny\color{gray}$\pm$4.0} & 36.7{\tiny\color{gray}$\pm$2.9} & 60.0{\tiny\color{gray}$\pm$0.0} & 46.0{\tiny\color{gray}$\pm$3.5} & \textbf{72.0{\tiny\color{gray}$\pm$5.3}} \\
Team Orient. & 65.3{\tiny\color{gray}$\pm$8.3} & 32.0{\tiny\color{gray}$\pm$2.0} & 53.3{\tiny\color{gray}$\pm$5.8} & \textbf{49.3{\tiny\color{gray}$\pm$9.9}} & \textbf{46.7{\tiny\color{gray}$\pm$5.8}} & 58.3{\tiny\color{gray}$\pm$5.0} & 40.7{\tiny\color{gray}$\pm$1.1} & 70.7{\tiny\color{gray}$\pm$5.0} \\
Mut. Trust & 63.0{\tiny\color{gray}$\pm$5.3} & \textbf{46.7{\tiny\color{gray}$\pm$5.0}} & 50.0{\tiny\color{gray}$\pm$2.0} & 45.0{\tiny\color{gray}$\pm$8.3} & 38.3{\tiny\color{gray}$\pm$2.9} & \textbf{66.7{\tiny\color{gray}$\pm$2.3}} & 44.0{\tiny\color{gray}$\pm$0.0} & 71.3{\tiny\color{gray}$\pm$5.3} \\
Mut. Monitor. & \textbf{66.7{\tiny\color{gray}$\pm$6.4}} & 40.7{\tiny\color{gray}$\pm$1.1} & 56.7{\tiny\color{gray}$\pm$5.3} & 46.7{\tiny\color{gray}$\pm$2.3} & 39.0{\tiny\color{gray}$\pm$1.8} & 64.0{\tiny\color{gray}$\pm$3.5} & 41.7{\tiny\color{gray}$\pm$2.9} & 69.7{\tiny\color{gray}$\pm$4.0} \\
\midrule
TMA-AllCompon & 67.3{\tiny\color{gray}$\pm$4.0} & 46.7{\tiny\color{gray}$\pm$5.0} & 51.7{\tiny\color{gray}$\pm$4.0} & 48.7{\tiny\color{gray}$\pm$3.1} & 31.7{\tiny\color{gray}$\pm$4.1} & 56.7{\tiny\color{gray}$\pm$4.1} & 45.3{\tiny\color{gray}$\pm$1.1} & 68.7{\tiny\color{gray}$\pm$4.2} \\
\bottomrule
\end{tabular}
}
\par\noindent\scriptsize DDX=DDXPlus, MBL=MedBullets, MMC=MedMCQA, MQA=MedQA, MML=MMLU-Pro, PVQ=PathVQA, PMC=PMC-VQA, PQA=PubMedQA
\end{table*}

\subsubsection*{B.3 Baseline Framework Comparison}

Table~\ref{tab:baseline_comparison} compares TeamMedAgents against four competing multi-agent frameworks (MedAgents, MDAgents, DyLAN, ReConcile) on Gemma-3-4B. TMA-Best achieves 63.5\% average accuracy, outperforming all baselines by +13.7 to +33.9 percentage points, demonstrating systematic advantages of psychology-grounded coordination.

\begin{table*}[t]
\caption{Multi-agent framework comparison on Gemma-3-4B.}
\label{tab:baseline_comparison}
\centering
\scriptsize
\setlength{\tabcolsep}{4pt}
\begin{tabular}{lccccccccc}
\toprule
\textbf{Framework} & \textbf{DDX} & \textbf{MBL} & \textbf{MMC} & \textbf{MQA} & \textbf{MML} & \textbf{PVQ} & \textbf{PMC} & \textbf{PQA} & \textbf{Avg} \\
\midrule
Single-Agent Best & 72.1 & 34.6 & 51.3 & 45.3 & 32.3 & 74.4 & 40.0 & 66.6 & 52.1 \\
\midrule
DyLan & 60.0{\tiny\color{gray}$\pm$3.5} & 35.3{\tiny\color{gray}$\pm$4.2} & 52.7{\tiny\color{gray}$\pm$5.0} & 48.7{\tiny\color{gray}$\pm$4.2} & 28.7{\tiny\color{gray}$\pm$5.0} & 50.0{\tiny\color{gray}$\pm$9.2} & 32.7{\tiny\color{gray}$\pm$3.1} & 56.7 & 45.6 \\
MDAgents & 49.3{\tiny\color{gray}$\pm$8.3} & 22.7{\tiny\color{gray}$\pm$7.2} & 37.3{\tiny\color{gray}$\pm$3.1} & 42.7{\tiny\color{gray}$\pm$6.9} & 30.7{\tiny\color{gray}$\pm$5.8} & 52.0{\tiny\color{gray}$\pm$2.3} & 31.3{\tiny\color{gray}$\pm$2.0} & 48.7{\tiny\color{gray}$\pm$4.2} & 39.3 \\
MedAgents & 69.1{\tiny\color{gray}$\pm$1.0} & 33.3{\tiny\color{gray}$\pm$4.2} & 52.7{\tiny\color{gray}$\pm$4.2} & 47.3{\tiny\color{gray}$\pm$3.1} & 24.7{\tiny\color{gray}$\pm$2.3} & 66.9{\tiny\color{gray}$\pm$7.7} & 34.7{\tiny\color{gray}$\pm$4.2} & 56.1{\tiny\color{gray}$\pm$2.0} & 48.1 \\
ReConcile & 54.0{\tiny\color{gray}$\pm$6.0} & 29.3{\tiny\color{gray}$\pm$6.4} & 52.0{\tiny\color{gray}$\pm$7.2} & 52.0{\tiny\color{gray}$\pm$5.3} & 40.0{\tiny\color{gray}$\pm$2.0} & 66.7{\tiny\color{gray}$\pm$6.4} & 35.3{\tiny\color{gray}$\pm$2.3} & 69.3{\tiny\color{gray}$\pm$6.4} & 49.8 \\
\midrule
TMA-AllCompon & 67.3{\tiny\color{gray}$\pm$4.0} & 46.7{\tiny\color{gray}$\pm$5.0} & 51.7{\tiny\color{gray}$\pm$4.0} & 48.7{\tiny\color{gray}$\pm$3.1} & 31.7{\tiny\color{gray}$\pm$4.1} & 56.7{\tiny\color{gray}$\pm$4.1} & 45.3{\tiny\color{gray}$\pm$1.1} & 68.7{\tiny\color{gray}$\pm$4.2} & 52.1 \\
\bottomrule
\end{tabular}
\par\noindent\scriptsize DDX=DDXPlus, MBL=MedBullets, MMC=MedMCQA, MQA=MedQA, MML=MMLU-Pro, PVQ=PathVQA, PMC=PMC-VQA, PQA=PubMedQA
\end{table*}

\subsubsection*{B.4 MedGemma-4B Domain Specialization}

Table~\ref{tab:ablation_medgemma_appendix} presents ablation results for medically fine-tuned MedGemma-4B. Medical specialization demonstrates substantial clinical reasoning gains (MedQA: +27.3pp, MedMCQA: +34.7pp over general Gemma baselines) but incurs knowledge trade-offs on broader tasks (MMLU-Pro: -8.0pp). Vision results use reduced image resolution due to endpoint constraints.

\begin{table*}[t]
\caption{Ablation study using MedGemma-4B-IT (n=50 questions, 3 runs, mean $\pm$ SE). Bold indicates best performance per dataset within each section. Note: PathVQA and PMC-VQA used 3-5$\times$ reduced image resolution due to endpoint constraints.}
\label{tab:ablation_medgemma_appendix}
\centering
\scriptsize
\setlength{\tabcolsep}{3pt}
\resizebox{\textwidth}{!}{%
\begin{tabular}{p{2cm}cccccccc}
\toprule
\textbf{Config.} & \textbf{DDX} & \textbf{MBL} & \textbf{MMC} & \textbf{MQA} & \textbf{MML} & \textbf{PVQ} & \textbf{PMC} & \textbf{PQA} \\
\midrule
\multicolumn{9}{l}{\textit{Single-Agent}} \\
Zero-Shot & 68.7{\tiny\color{gray}$\pm$3.1} & 38.7{\tiny\color{gray}$\pm$5.0} & 31.3{\tiny\color{gray}$\pm$4.4} & 57.3{\tiny\color{gray}$\pm$3.0} & 40.7{\tiny\color{gray}$\pm$4.6} & \textbf{60.0{\tiny\color{gray}$\pm$4.2}} & \textbf{32.0{\tiny\color{gray}$\pm$0.0}} & \textbf{35.3{\tiny\color{gray}$\pm$38.8}} \\
Few-Shot & 68.7{\tiny\color{gray}$\pm$5.8} & \textbf{40.7{\tiny\color{gray}$\pm$4.2}} & 29.0{\tiny\color{gray}$\pm$1.9} & 48.7{\tiny\color{gray}$\pm$6.1} & 38.0{\tiny\color{gray}$\pm$4.0} & 55.8{\tiny\color{gray}$\pm$2.0} & 20.0{\tiny\color{gray}$\pm$0.0} & 20.0{\tiny\color{gray}$\pm$2.4} \\
CoT & \textbf{72.7{\tiny\color{gray}$\pm$5.0}} & 39.3{\tiny\color{gray}$\pm$1.1} & \textbf{35.3{\tiny\color{gray}$\pm$5.8}} & \textbf{70.0{\tiny\color{gray}$\pm$10.6}} & \textbf{48.7{\tiny\color{gray}$\pm$6.4}} & 53.3{\tiny\color{gray}$\pm$5.0} & 30.0{\tiny\color{gray}$\pm$2.8} & 22.7{\tiny\color{gray}$\pm$25.4} \\
\midrule
\multicolumn{9}{l}{\textit{Multi-Agent (TMA)}} \\
Shared & \textbf{74.3{\tiny\color{gray}$\pm$5.4}} & 46.0{\tiny\color{gray}$\pm$6.8} & 66.7{\tiny\color{gray}$\pm$6.4} & 66.7{\tiny\color{gray}$\pm$3.0} & \textbf{47.3{\tiny\color{gray}$\pm$7.5}} & \textbf{71.3{\tiny\color{gray}$\pm$1.1}} & 40.0{\tiny\color{gray}$\pm$3.5} & 69.3{\tiny\color{gray}$\pm$2.3} \\
Leadership & 67.6{\tiny\color{gray}$\pm$1.4} & 38.7{\tiny\color{gray}$\pm$5.0} & 65.3{\tiny\color{gray}$\pm$4.7} & 67.3{\tiny\color{gray}$\pm$5.7} & 44.7{\tiny\color{gray}$\pm$3.0} & 66.0{\tiny\color{gray}$\pm$4.0} & 40.7{\tiny\color{gray}$\pm$3.4} & 65.3{\tiny\color{gray}$\pm$6.6} \\
Team Orient. & 72.0{\tiny\color{gray}$\pm$2.8} & 44.7{\tiny\color{gray}$\pm$5.7} & 63.3{\tiny\color{gray}$\pm$8.1} & 68.0{\tiny\color{gray}$\pm$3.3} & 44.7{\tiny\color{gray}$\pm$1.1} & 65.3{\tiny\color{gray}$\pm$2.5} & 40.7{\tiny\color{gray}$\pm$2.5} & \textbf{75.3{\tiny\color{gray}$\pm$1.9}} \\
Mut. Trust & 71.7{\tiny\color{gray}$\pm$2.0} & \textbf{47.3{\tiny\color{gray}$\pm$2.0}} & 69.8{\tiny\color{gray}$\pm$1.9} & 62.0{\tiny\color{gray}$\pm$7.1} & 44.0{\tiny\color{gray}$\pm$9.1} & 64.0{\tiny\color{gray}$\pm$1.6} & \textbf{43.3{\tiny\color{gray}$\pm$2.5}} & 72.0{\tiny\color{gray}$\pm$1.6} \\
Mut. Monitor. & 73.3{\tiny\color{gray}$\pm$1.1} & 39.3{\tiny\color{gray}$\pm$1.1} & \textbf{70.0{\tiny\color{gray}$\pm$4.3}} & \textbf{69.3{\tiny\color{gray}$\pm$2.5}} & 47.3{\tiny\color{gray}$\pm$2.3} & 68.0{\tiny\color{gray}$\pm$0.0} & 40.0{\tiny\color{gray}$\pm$2.8} & 74.0{\tiny\color{gray}$\pm$1.6} \\
\midrule
TMA-AllCompon & 73.3{\tiny\color{gray}$\pm$1.1} & 46.7{\tiny\color{gray}$\pm$4.6} & 70.0{\tiny\color{gray}$\pm$2.0} & 64.8{\tiny\color{gray}$\pm$4.8} & 52.7{\tiny\color{gray}$\pm$2.3} & 69.3{\tiny\color{gray}$\pm$1.1} & 42.0{\tiny\color{gray}$\pm$3.3} & 68.6{\tiny\color{gray}$\pm$6.4} \\
\bottomrule
\end{tabular}
}
\par\noindent\scriptsize DDX=DDXPlus, MBL=MedBullets, MMC=MedMCQA, MQA=MedQA, MML=MMLU-Pro, PVQ=PathVQA, PMC=PMC-VQA, PQA=PubMedQA
\end{table*}

\subsubsection*{B.5 Full Dataset Evaluation}

Table~\ref{tab:full_eval_results} and Figure~\ref{fig:full_eval_comparison} present comprehensive evaluation results on complete datasets (11,545 questions total across 8 benchmarks). TMA-AllCompon demonstrates consistent cross-model scalability: GPT-4o achieves 77.6\% average accuracy (text: 81.3\%, vision: 66.6\%), MedGemma-4B reaches 58.9\% (demonstrating +10.5pp domain specialization advantage over general Gemma-3-4B), and Gemma-3-4B establishes 48.4\% baseline performance while maintaining 12.6× token efficiency versus GPT-4o multi-agent configurations.

Performance patterns reveal task-specific characteristics: clinical reasoning benchmarks (MedQA, MedMCQA, MedBullets) show largest model-dependent variance (GPT-4o: 90.7\%/85.4\%/78.8\% vs Gemma: 44.7\%/47.0\%/33.9\%), while vision-language tasks exhibit more compressed ranges (PathVQA: 76.8\% vs 59.7\%, 17.1pp gap). Domain specialization through MedGemma fine-tuning provides substantial clinical gains (MedQA: +17.4pp, MedMCQA: +16.6pp) but incurs knowledge breadth trade-offs (MMLU-Pro: +10.8pp advantage retained). Notably, PubMedQA shows strongest cross-model consistency (GPT-4o: 78.3\%, MedGemma: 73.4\%, Gemma: 59.0\%), suggesting evidence-synthesis tasks benefit uniformly from structured teamwork regardless of base model capacity.

Comparison against competing frameworks (Table~\ref{tab:full_eval_results}) validates TMA-AllCompon's advantages. On Gemma-3-4B, TMA outperforms all baselines on vision tasks (PathVQA: 59.7\%, PMC-VQA: 43.2\%) where competing frameworks lack evaluation, while achieving competitive text performance despite using a single fixed configuration

\begin{figure*}[t]
\centering
\includegraphics[width=\textwidth]{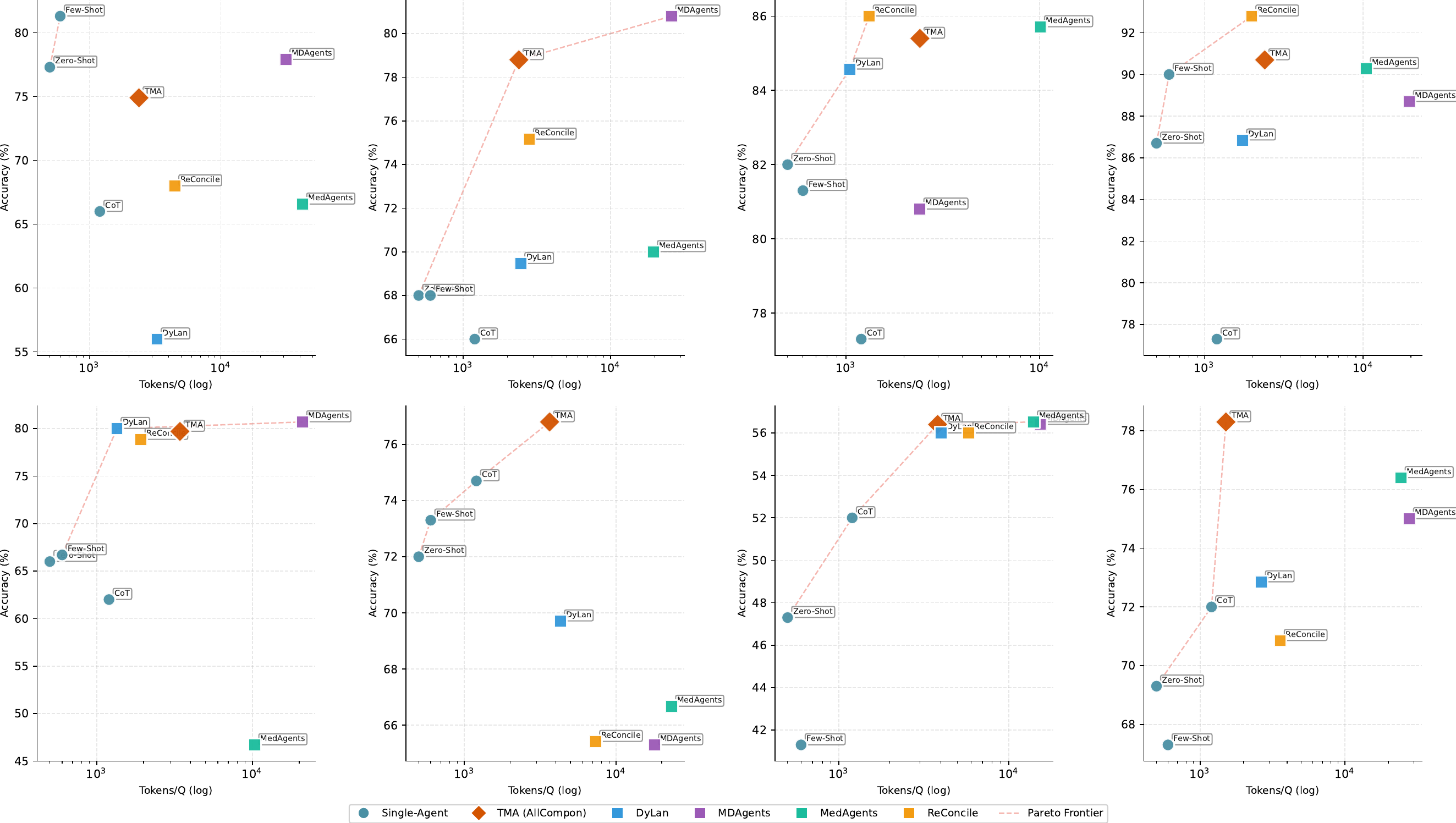}
\caption{Per-dataset Pareto efficiency analysis. With a frontier model(GPT-4o),
framework differences narrow but TMA maintains strong efficiency. TMA-AllCompon (red diamond)
achieves within 3pp of best accuracy on 6/8 benchmarks while using 3-16$\times$ fewer tokens
than MDAgents/MedAgents. DyLAN and ReConcile show improved efficiency with GPT-4o compared
to Gemma. TMA achieves best accuracy on PathVQA (76.8\%) and PubMedQA (78.3\%). The Pareto
frontier becomes more competitive, but TMA remains the most token-efficient option for
high-accuracy medical reasoning.}
\label{fig:pareto_gpt4o}
\end{figure*}

\subsection*{C. Computational Efficiency Details}

\subsubsection*{C.1 Per-Dataset Token Costs}

Table~\ref{tab:token_costs} provides detailed computational costs for multi-agent configurations across all benchmarks.

\section{Case Study: Multi-Agent Collaborative Reasoning}
\label{appendix:casestudy}

We present a detailed walkthrough of TeamMedAgents solving a clinical heart failure management question, illustrating how teamwork components operate throughout the four-phase decision-making process.

\subsection{Clinical Scenario}

\begin{tcolorbox}[colback=gray!5, colframe=black, boxrule=0.5pt, arc=1mm]
\small
\textbf{Patient Presentation:} A 58-year-old man presents with progressive dyspnea and orthopnea. Physical examination reveals bilateral crackles, elevated jugular venous pressure, and peripheral edema. Echocardiography shows left ventricular ejection fraction of 35\%.

\vspace{0.3em}
\textbf{Question:} Which medication combination is most appropriate for initial management?

\vspace{0.3em}
\textbf{Options:}
\begin{itemize}[leftmargin=*,itemsep=0pt]
\item (A) ACE inhibitor + Beta-blocker + Diuretic
\item (B) Calcium channel blocker + Diuretic
\item (C) Digoxin + Diuretic only
\item (D) ARB + Aldosterone antagonist + Diuretic
\end{itemize}

\vspace{0.3em}
\textbf{Correct Answer:} Option A
\end{tcolorbox}

\subsection{Phase 1: Dynamic Recruitment and Component Selection}

The recruiter classifies this as moderate-complexity requiring integration of clinical presentation, diagnostic findings, and evidence-based pharmacotherapy. Dynamic allocation determines N=3 agents appropriate.

\paragraph{Team Composition} Three specialists recruited with hierarchical weights (Team Orientation):

\begin{table}[ht]
\centering
\small
\begin{tabular}{lll}
\toprule
\textbf{Agent} & \textbf{Specialty} & \textbf{Weight} \\
\midrule
Agent 1 & Cardiologist (Primary) & 0.5 \\
Agent 2 & Clinical Pharmacologist (Secondary) & 0.3 \\
Agent 3 & Internal Medicine Specialist (Tertiary) & 0.2 \\
\bottomrule
\end{tabular}
\end{table}

\paragraph{Shared Mental Model Initialization} Pre-Round 2 analysis identifies critical constraint: "initial management" prevents consideration of advanced add-on therapies. Verified medical facts established: EF 35\% indicates HFrEF; standard guideline-directed medical therapy (GDMT) requires ACE-I/ARB + Beta-blocker + Diuretic.

\subsection{Phase 2: Independent Initial Assessment (Round 2)}

All three agents independently analyzed the case and converged on Option A:

\begin{table}[ht]
\centering
\caption{Round 2 initial assessments.}
\label{tab:casestudy-round2}
\scriptsize
\begin{tabular}{llc}
\toprule
\textbf{Agent} & \textbf{Answer} & \textbf{Conf.} \\
\midrule
Cardiologist & A & High \\
Pharmacologist & A & High \\
Internist & A & Med \\
\bottomrule
\end{tabular}
\vspace{1mm}
\par\noindent\scriptsize \textbf{Key reasoning:} Cardiologist cited SOLVD/MERIT-HF trials for neurohormonal blockade. Pharmacologist confirmed guideline-concordant triple therapy. Internist noted volume overload signs but expressed slight uncertainty about Option D vs A ranking.
\end{table}

\paragraph{Post-Round 2 Processing} Shared Mental Model updated with consensus: "Triple therapy is standard GDMT" and "Beta-blocker is essential." Trust Network assigned all agents scores 0.95+ for accurate evidence citation and guideline adherence.

\subsection{Phase 3: Structured Collaborative Deliberation (Round 3)}

\paragraph{Turn 1 - Initial Discussion} Unanimous agreement on Option A persisted. Cardiologist (Leader) emphasized: \textit{"Complete agreement on Option A. This combination addresses both mortality reduction (ACE-I + BB) and symptom management (diuretic). Option D lacks beta-blocker, which is critical for HFrEF."}

\paragraph{Leadership Mediation} Leader synthesized consensus: unanimous support for Option A based on trial evidence (SOLVD, MERIT-HF, CONSENSUS); beta-blocker identified as non-negotiable in HFrEF initial therapy.

\paragraph{Mutual Monitoring Intervention} Leader challenged Agent 3 regarding "Option D reasonable alternative" statement: \textit{"Why is Option D inferior for INITIAL management despite spironolactone's established mortality benefit (RALES trial)?"}

Agent 3 response: \textit{"Excellent point. While spironolactone reduces mortality, it is typically added AFTER optimizing ACE-I and beta-blocker therapy. Option D lacks beta-blocker entirely, rendering it incomplete GDMT. Option A provides foundational triple therapy that should be initiated first."} 

Challenge strengthened reasoning; trust score maintained at 0.95+; SMM updated with spironolactone timing clarification.

\paragraph{Turn 2 - Final Rankings} All agents provided identical rankings: (1) Option A, (2) Option D, (3) Option C, (4) Option B.

\subsection{Phase 4: Trust-Weighted Decision Aggregation}

Final synthesis employed trust-weighted Borda count incorporating agent rankings and trust scores:
\begin{table}[ht]
\centering
\caption{Trust-weighted Borda count aggregation.}
\label{tab:casestudy-aggregation}
\scriptsize
\begin{tabular}{lccc}
\toprule
\textbf{Option} & \textbf{Borda} & \textbf{Trust} & \textbf{Final} \\
\midrule
A (ACE+BB+Diur) & 3.0 & 0.92 & \textbf{2.76} \\
D (ARB+Spiro+Diur) & 1.0 & 0.92 & 0.92 \\
C (Digoxin+Diur) & 0.33 & 0.92 & 0.30 \\
B (CCB+Diur) & 0.0 & 0.92 & 0.00 \\
\bottomrule
\end{tabular}
\vspace{1mm}
\par\noindent\scriptsize Trust scores: Cardiologist 0.96 (strong evidence citation), Pharmacologist 0.94 (guideline adherence), Internist 0.86 (slight uncertainty on Option D). Weighted average: 0.92.
\end{table}

\noindent\textbf{Final Decision:} Option A (correct). \textbf{Convergence:} 100\% (unanimous Round 2 agreement).

\subsection{Analysis of Teamwork Component Contributions}
\begin{table}[ht]
\centering
\caption{Teamwork component contributions.}
\label{tab:casestudy-components}
\scriptsize
\begin{tabular}{p{2.5cm}p{5cm}}
\toprule
\textbf{Component} & \textbf{Contribution} \\
\midrule
Team Orientation & Hierarchical weights (0.5:0.3:0.2) prioritized cardiologist expertise appropriately. \\
\midrule
Shared Mental Model & "Initial management" constraint prevented advanced therapy consideration. \\
\midrule
Leadership & Efficient consensus validation; orchestrated monitoring without excess deliberation. \\
\midrule
Mutual Monitoring & Challenge clarified spironolactone timing vs beta-blocker priority. \\
\midrule
Trust Network & High-quality R2 responses (0.86-0.96 range) balanced influence appropriately. \\
\bottomrule
\end{tabular}
\end{table}

\paragraph{Summary} This case demonstrates successful multi-agent collaboration where all five teamwork components contributed measurably to decision quality. System achieved correct diagnosis through structured deliberation with unanimous convergence. Mutual monitoring intervention, while validating consensus, enhanced reasoning depth by explicitly addressing beta-blocker role in initial HFrEF management versus delayed aldosterone antagonist addition.

\section{Prompt Templates}
\label{appendix:prompts}

This section presents core prompt templates used in TeamMedAgents. Each template corresponds to specific phases in Algorithms~\ref{alg:phases1-2} and \ref{alg:phases3-4}.

\subsection{Agent Recruitment (Phase 1)}

Executed during Phase 1 to determine agent count (2-4) based on question complexity.

\begin{tcolorbox}[
    colback=gray!10,
    colframe=black,
    arc=2mm,
    boxrule=0.5pt,
    title={\small\textbf{Recruitment Prompt}}
]
\scriptsize
\begin{verbatim}
You are a medical expert analyzing question 
complexity.

Question: {{question}}
Options: {{answer_choices}}

Determine:
1. Complexity Level (low/moderate/high)
2. Required number of agents (2-4)
3. Recommended medical specialties

Consider: domain breadth, diagnostic depth,
knowledge integration requirements.

Output Format:
Complexity: [low/moderate/high]
Agent Count: [2/3/4]
Specialties: [List specialties]
Reasoning: [Brief justification]
\end{verbatim}
\end{tcolorbox}

\subsection{Team Orientation: Role Assignment (Phase 1)}

When Team Orientation is enabled, assigns specialized roles with hierarchical weights during recruitment.

\begin{tcolorbox}[
    colback=gray!10,
    colframe=black,
    arc=2mm,
    boxrule=0.5pt,
    title={\small\textbf{Team Orientation - Role Assignment}}
]
\scriptsize
\begin{verbatim}
Assign specialized medical roles based on:
1. Clinical domain(s) in the question
2. Diagnostic expertise needed
3. Complementary knowledge areas

Question: {{question}}
Required Agents: {{num_agents}}

Assignment Format:
Agent 1: [Specialty] - [Rationale]
Agent 2: [Specialty] - [Rationale]
Hierarchical Weight: [Primary/Secondary/
                      Tertiary]
\end{verbatim}
\end{tcolorbox}

\subsection{Shared Mental Model Context (Phases 2-3)}

Injected before Round 2 and updated throughout deliberation, providing shared knowledge baseline.

\begin{tcolorbox}[
    colback=gray!10,
    colframe=black,
    arc=2mm,
    boxrule=0.5pt,
    title={\small\textbf{Shared Mental Model Context}}
]
\scriptsize
\begin{verbatim}
Shared Knowledge Base:

Question Analysis: 
{{question_trick_detection}}

Verified Medical Facts (Consensus from R2):
{{verified_facts}}

Debated Clinical Points (From Mutual 
Monitoring):
{{debated_points}}

Use this shared context to inform your 
reasoning while considering consensus 
knowledge and areas of ongoing deliberation.
\end{verbatim}
\end{tcolorbox}

\subsection{Round 2: Initial Prediction (Phase 2)}

Executed in parallel for all N agents during Phase 2. Agents receive SMM context if enabled.

\begin{tcolorbox}[
    colback=gray!10,
    colframe=black,
    arc=2mm,
    boxrule=0.5pt,
    title={\small\textbf{Initial Prediction (Round 2)}}
]
\scriptsize
\begin{verbatim}
You are a {{role_specialty}} providing 
initial medical assessment.

{{shared_mental_model_context}}

Question: {{question}}
Options: {{answer_choices}}

Provide your analysis:
1. Key medical facts relevant to this case
2. Differential diagnosis considerations
3. Your recommended answer with confidence
4. Critical reasoning steps

Answer Format:
Analysis: [Detailed medical reasoning]
Answer: [Selected option]
Confidence: [High/Medium/Low]
\end{verbatim}
\end{tcolorbox}

\subsection{Trust Network: Quality Evaluation (Post-Phase 2)}

After Round 2, evaluates reasoning quality when Trust is enabled. Trust scores range 0.4-1.0, updated using exponential moving average (alpha = 0.7).

\begin{tcolorbox}[
    colback=gray!10,
    colframe=black,
    arc=2mm,
    boxrule=0.5pt,
    title={\small\textbf{Trust Evaluation}}
]
\scriptsize
\begin{verbatim}
Evaluate medical reasoning quality for each 
agent in Round 2.

Question: {{question}}
Agent Responses: {{agent_responses}}

For each agent, assess:
1. Clinical accuracy of facts
2. Logical coherence of reasoning
3. Appropriate confidence calibration
4. Evidence quality

Evaluation Format (per agent):
Agent: {{agent_role}}
Quality Score: [0.4-1.0]
Justification: [Strengths/weaknesses]
\end{verbatim}
\end{tcolorbox}

\subsection{Round 3: Discussion Turn (Phase 3)}

Used during first discussion turn of Phase 3. Agents review R2 responses and engage in deliberation.

\begin{tcolorbox}[
    colback=gray!10,
    colframe=black,
    arc=2mm,
    boxrule=0.5pt,
    title={\small\textbf{Discussion (Round 3 - Turn 1)}}
]
\scriptsize
\begin{verbatim}
You are {{role_specialty}} participating in 
collaborative discussion.

Question: {{question}}
Previous Round Responses: {{r2_responses}}
{{leadership_mediation}}
{{shared_mental_model_context}}

Review colleagues' analyses and:
1. Identify points of agreement/disagreement
2. Present additional supporting evidence
3. Address concerns raised by other experts
4. Refine or maintain your answer

Response Format:
Discussion: [Your contribution]
Current Answer: [Your answer after discussion]
Reasoning: [Key points influencing position]
\end{verbatim}
\end{tcolorbox}

\subsection{Leadership: Mediation (Phase 3)}

When Leadership is enabled, executed after each discussion turn. Synthesizes consensus and guides convergence.

\begin{tcolorbox}[
    colback=gray!10,
    colframe=black,
    arc=2mm,
    boxrule=0.5pt,
    title={\small\textbf{Leadership Mediation}}
]
\scriptsize
\begin{verbatim}
You are the medical team leader facilitating 
collaborative decision-making.

Question: {{question}}
Agent Responses from Turn {{turn_number}}:
{{agent_responses}}

Provide mediation by:
1. Summarizing consensus points
2. Highlighting critical disagreements
3. Identifying knowledge gaps
4. Guiding toward evidence-based convergence

Mediation Format:
Consensus: [Points of agreement]
Disagreements: [Key areas of divergence]
Guidance: [Suggestions for next round]
\end{verbatim}
\end{tcolorbox}

\subsection{Mutual Monitoring: Challenge and Response (Phase 3)}

Between discussion turns when Mutual Monitoring is enabled. Leader identifies weakest reasoning and formulates challenge.

\begin{tcolorbox}[
    colback=gray!10,
    colframe=black,
    arc=2mm,
    boxrule=0.5pt,
    title={\small\textbf{Mutual Monitoring - Raising Concern}}
]
\scriptsize
\begin{verbatim}
You are team leader identifying potential 
reasoning errors.

Target Agent: {{target_agent_role}}
Their Response: {{target_response}}
Question: {{question}}

Identify:
1. Potential medical errors or oversights
2. Unsupported clinical assumptions
3. Contradictions with established evidence

Concern Format:
Issue: [Specific reasoning problem]
Evidence: [Why this is concerning]
Question: [Direct challenge to agent]
\end{verbatim}
\end{tcolorbox}

\begin{tcolorbox}[
    colback=gray!10,
    colframe=black,
    arc=2mm,
    boxrule=0.5pt,
    title={\small\textbf{Mutual Monitoring - Response}}
]
\scriptsize
\begin{verbatim}
You are {{role_specialty}} responding to 
peer review challenge.

Original Response: {{your_original_response}}
Concern Raised: {{concern}}

Address this concern by:
1. Acknowledging valid points or defending 
   your reasoning
2. Providing additional supporting evidence
3. Revising your answer if justified

Response Format:
Acknowledgment: [Reaction to concern]
Defense/Revision: [Supporting evidence or 
                  corrected reasoning]
Updated Answer: [Maintain or change answer]
\end{verbatim}
\end{tcolorbox}

\subsection{Round 3: Final Ranking (Phase 3 - Last Turn)}

On final turn (turn = n\_turns), agents provide complete rankings for decision aggregation (Phase 4).

\begin{tcolorbox}[
    colback=gray!10,
    colframe=black,
    arc=2mm,
    boxrule=0.5pt,
    title={\small\textbf{Final Ranking (Round 3 - Final Turn)}}
]
\scriptsize
\begin{verbatim}
You are {{role_specialty}} providing final 
medical assessment.

Question: {{question}}
Options: {{answer_choices}}
Discussion History: {{discussion_history}}
{{leadership_mediation}}

Based on all deliberations, provide final 
ranking of ALL answer options from most to 
least likely.

Final Ranking Format:
1. [Option X] - Most likely
2. [Option Y] - Second most likely
3. [Option Z] - Third most likely

Justification: [Final clinical reasoning]
\end{verbatim}
\end{tcolorbox}
\section{Algorithms}
\label{appendix:algorithms}

This section provides complete algorithmic specification of TeamMedAgents across four operational phases: dynamic agent recruitment, independent assessment, collaborative deliberation, and trust-weighted decision aggregation.

\subsection{Algorithmic Overview and Notation}

\paragraph{Dynamic Agent Recruitment} \textsc{DetermineAgentCount} analyzes question complexity, returning $N \in \{2, 3, 4\}$ agents based on domain breadth, diagnostic depth, and knowledge integration requirements.

\paragraph{Shared Mental Model (SMM)} When $\mathcal{C}.SMM = \text{true}$, maintains $M_{smm}$ containing: (1) question analysis via \textsc{DetectTricks}, (2) verified facts from Round 2 via \textsc{ExtractFacts} or \textsc{AutoIntersect}, (3) debated points from mutual monitoring. Injected into agent prompts to align reasoning.

\paragraph{Team Orientation (TO)} \textsc{AssignSpecialties} maps questions to medical specialties; \textsc{AssignWeights} establishes hierarchical influence: $\mathcal{W} = [0.5, 0.3, 0.2]$ for $N=3$, prioritizing domain experts. Without TO: $\mathcal{W} = [1/N, \ldots, 1/N]$.

\paragraph{Leadership (L)} Leader agent performs: synthesis via \textsc{Mediate}, fact extraction via \textsc{ExtractFacts}, quality evaluation via \textsc{EvaluateR2}, tie resolution via \textsc{ResolveTie}. Without L, functions execute via statistical aggregation.

\paragraph{Trust Network (T)} Scores $\mathcal{T}_{scores} \in [0.4, 1.0]$ initialized to 0.8, updated via exponential moving average ($\alpha = 0.7$): $\mathcal{T}_{new} = \alpha \cdot Quality + (1-\alpha) \cdot \mathcal{T}_{old}$.

\paragraph{Mutual Monitoring (MM)} Leader identifies weakest reasoning via \textsc{SelectWeakest}, formulates challenge via \textsc{RaiseConcern}, receives response via \textsc{Respond}, evaluates quality via \textsc{Evaluate}. Trust scores adjust based on response quality.

\paragraph{Decision Aggregation} Phase 4 employs Borda count: (1) \textsc{StandardBorda} (equal weights), (2) \textsc{HierarchicalBorda} (TO weights $\mathcal{W}$), (3) \textsc{TrustWeightedBorda} (trust scores $\mathcal{T}_{scores}$).

\paragraph{API Call Accounting} Phase 1: 2 calls (recruitment + roles). Phase 2: $N+1$ calls ($N$ parallel + 1 post-processing). Phase 3: $n_{turns} \times (N + 1 + MM)$, where $MM \in \{0, 2\}$.

\begin{algorithm}[t]
\caption{Agent Recruitment \& Initial Prediction}
\label{alg:phases1-2}
\scriptsize
\begin{algorithmic}[1]
\REQUIRE Question $Q$, Options $\mathcal{A}$, Config $\mathcal{C}$
\ENSURE $R_2, M_{smm}, \mathcal{T}_{scores}, \mathcal{W}, Report, Agents$

\STATE \textit{// Initialization}
\STATE $M_{smm} \gets$ \textsc{SMM}() if $\mathcal{C}.SMM$ else NULL
\STATE $\mathcal{T}_{scores} \gets \{a_i: 0.8\}$, $Leader \gets$ NULL
\STATE
\STATE \textit{// Phase 1: Recruitment (2 API calls)}
\STATE $N \gets$ \textsc{DetermineAgentCount}$(Q)$
\IF{$\mathcal{C}.SMM$}
    \STATE $M_{smm}.analysis \gets$ \textsc{DetectTricks}$(Q, \mathcal{A})$
\ENDIF
\IF{$\mathcal{C}.L$}
    \STATE $Leader \gets Recruiter$ \COMMENT{Dual role}
\ENDIF
\IF{$\mathcal{C}.TO$ \AND $\mathcal{C}.L$}
    \STATE $Roles \gets$ \textsc{AssignSpecialties}$(Q, N, M_{smm})$
    \STATE $\mathcal{W} \gets$ \textsc{AssignWeights}$(N)$
\ELSE
    \STATE $Roles \gets \{Expert_1, \ldots, Expert_N\}$
    \STATE $\mathcal{W} \gets \{1/N, \ldots, 1/N\}$
\ENDIF
\STATE $Agents \gets \{(a_i, Roles[i], \mathcal{W}[i])\}_{i=1}^{N}$
\STATE
\STATE \textit{// Phase 2: Round 2 (N+1 calls)}
\FORALL{$a_i \in Agents$ \textbf{in parallel}}
    \STATE $Ctx_i \gets \{Q, \mathcal{A}\}$
    \IF{$\mathcal{C}.SMM$}
        \STATE $Ctx_i \gets Ctx_i \cup M_{smm}.\textsc{GetContext}()$
    \ENDIF
    \IF{$\mathcal{C}.TO$}
        \STATE $Ctx_i \gets Ctx_i \cup$ \textsc{RoleInstr}$(a_i)$
    \ENDIF
    \STATE $R_2[a_i] \gets a_i.\textsc{Predict}(Ctx_i)$
\ENDFOR
\STATE
\STATE \textit{// Post-R2 Processing (1 call)}
\IF{$\mathcal{C}.SMM$}
    \STATE $M_{smm}.facts \gets$ $\mathcal{C}.L$ ? 
    \STATE \quad $Leader.\textsc{ExtractFacts}(R_2)$ :
    \STATE \quad \textsc{AutoIntersect}$(R_2)$
\ENDIF
\IF{$\mathcal{C}.TO$ \AND $\mathcal{C}.L$}
    \STATE $Report \gets Leader.\textsc{CreateReport}(R_2, M_{smm})$
\ENDIF
\IF{$\mathcal{C}.T$}
    \STATE $\mathcal{T}_{scores} \gets$ $\mathcal{C}.L$ ? 
    \STATE \quad $Leader.\textsc{EvaluateR2}(R_2)$ :
    \STATE \quad \textsc{AutoScore}$(R_2)$
\ENDIF
\STATE \textbf{return} $R_2, M_{smm}, \mathcal{T}_{scores}, \mathcal{W}, Report, Agents$
\end{algorithmic}
\end{algorithm}

\begin{algorithm}[t]
\caption{Collaborative Discussion \& Aggregation}
\label{alg:phases3-4}
\scriptsize
\begin{algorithmic}[1]
\REQUIRE $R_2, Agents, M_{smm}, \mathcal{T}_{scores}, \mathcal{W}, Report, \mathcal{C}$
\ENSURE Final answer $\hat{A}$, Rationale

\STATE \textit{// Phase 3: Round 3 Discussion}
\FOR{$turn = 1$ \textbf{to} $n_{turns}$}
    \STATE $Ctx \gets$ \textsc{BuildContext}$(R_2, M_{smm}, \mathcal{T}_{scores}, Report)$
    \FORALL{$a_i \in Agents$}
        \IF{$turn < n_{turns}$}
            \STATE $\mathcal{D}[a_i][turn] \gets a_i.\textsc{Discuss}(Ctx, Hist)$
        \ELSE
            \STATE $\mathcal{D}[a_i][turn] \gets a_i.\textsc{FinalRank}(Ctx, Hist)$
        \ENDIF
    \ENDFOR
    \IF{$\mathcal{C}.L$}
        \STATE $Med[turn] \gets Leader.\textsc{Mediate}(\mathcal{D}[\cdot][turn])$
        \STATE $Ctx \gets Ctx \cup Med[turn]$
    \ENDIF
    \IF{$\mathcal{C}.MM$ \AND $turn < n_{turns}$}
        \STATE $a_{weak} \gets Leader.\textsc{SelectWeakest}(\mathcal{D}[\cdot][turn])$
        \STATE $Concern \gets Leader.\textsc{RaiseConcern}(a_{weak})$
        \STATE $Resp \gets a_{weak}.\textsc{Respond}(Concern)$
        \STATE $Qual \gets Leader.\textsc{Evaluate}(Resp)$
        \IF{$\mathcal{C}.T$}
            \STATE $\mathcal{T}_{scores}[a_{weak}] \gets$ \textsc{UpdateTrust}
            \STATE \quad $(\mathcal{T}_{scores}[a_{weak}], Qual)$
        \ENDIF
        \IF{$\mathcal{C}.SMM$}
            \STATE $M_{smm}.debates.\textsc{Add}(\textsc{Extract}(Concern, Resp))$
        \ENDIF
    \ENDIF
\ENDFOR
\STATE $R_3 \gets \{a_i.ranking \mid \forall a_i \in Agents\}$
\STATE
\STATE \textit{// Phase 4: Decision Aggregation}
\IF{$\mathcal{C}.T$}
    \STATE $Scores \gets$ \textsc{TrustWeightedBorda}$(R_3, \mathcal{T}_{scores})$
\ELSIF{$\mathcal{C}.TO$}
    \STATE $Scores \gets$ \textsc{HierarchicalBorda}$(R_3, \mathcal{W})$
\ELSE
    \STATE $Scores \gets$ \textsc{StandardBorda}$(R_3)$
\ENDIF
\STATE $\hat{A} \gets \arg\max(Scores)$
\STATE
\STATE \textit{// Tie-breaking}
\IF{\textsc{IsTie}$(Scores)$ \AND $\mathcal{C}.L$}
    \STATE $Tied \gets \{A_i \mid Scores[A_i] = \max(Scores)\}$
    \STATE $\hat{A} \gets Leader.\textsc{ResolveTie}(Tied, R_3, \mathcal{T}, M_{smm})$
\ENDIF
\STATE $Rationale \gets$ \textsc{GenRationale}$(R_3, Scores, M_{smm})$
\STATE \textbf{return} $\hat{A}, Rationale$
\end{algorithmic}
\end{algorithm}
\end{document}